\documentclass[10pt,twocolumn,letterpaper]{article}
\usepackage[accsupp]{axessibility} 
\usepackage{algorithm}
\usepackage{amssymb}
\usepackage{amsmath}
\usepackage{listings}
\usepackage{graphicx}  
\usepackage{booktabs}  
\usepackage{wrapfig}
\usepackage{colortbl}
\usepackage[makeroom]{cancel}
\usepackage{cases}
\definecolor{gray}{rgb}{0.5,0.5,0.5}
\definecolor{darkergreen}{RGB}{21, 152, 56}
\definecolor{RoyalBlue}{RGB}{65,105,225}
\definecolor{YellowOrange}{RGB}{255,165,0}
\definecolor{gray94}{gray}{.92}
\definecolor{gray90}{gray}{.90}
\definecolor{gray85}{gray}{.85}
\usepackage{color,soul} 
\sethlcolor{gray90} 

\usepackage{pifont}%
\usepackage{xspace}
\usepackage{placeins}
\usepackage{etoolbox}
\makeatletter
\AfterEndEnvironment{algorithm}{\let\@algcomment\relax}
\AtEndEnvironment{algorithm}{\kern2pt\hrule\relax\vskip3pt\@algcomment}
\let\@algcomment\relax
\newcommand\algcomment[1]{\def\@algcomment{\footnotesize#1}}
\renewcommand\fs@ruled{\def\@fs@cfont{\bfseries}\let\@fs@capt\floatc@ruled
  \def\@fs@pre{\hrule height.8pt depth0pt \kern2pt}%
  \def\@fs@post{}%
  \def\@fs@mid{\kern2pt\hrule\kern2pt}%
  \let\@fs@iftopcapt\iftrue}
\makeatother

\usepackage{iccv}              

\definecolor{iccvblue}{rgb}{0.21,0.49,0.74}
\usepackage[pagebackref,breaklinks,colorlinks,allcolors=iccvblue]{hyperref}

\def\paperID{7332} 
\def\confName{ICCV}
\def\confYear{2025}

\title{Met$^2$Net: A Decoupled Two-Stage Spatio-Temporal Forecasting Model for Complex Meteorological Systems}

\author{
\bf{Shaohan Li}$^{1,2*}$~~~~Hao Yang$^{1*}$~~~~Min Chen$^{1,2,3\dagger}$~~~~Xiaolin Qin$^{2,3\dagger}$\\
$^1$Chengdu University of Information Technology\\
$^2$Chengdu Institute of Computer Applications, Chinese Academy of Sciences\\
$^3$University of Chinese Academy of Sciences\\
}

\begin{document}
\maketitle
\begin{abstract}

The increasing frequency of extreme weather events due to global climate change urges accurate weather prediction. 
Recently, great advances have been made by the \textbf{end-to-end methods}, thanks to deep learning techniques, but they face limitations of \textit{representation inconsistency} in multivariable integration and struggle to effectively capture the dependency between variables, which is required in complex weather systems.
Treating different variables as distinct modalities and applying a \textbf{two-stage training approach} from multimodal models can partially alleviate this issue, but due to the inconformity in training tasks between the two stages, the results are often suboptimal. To address these challenges, we propose an implicit two-stage training method, configuring separate encoders and decoders for each variable. In detailed, in the first stage, the Translator is frozen while the Encoders and Decoders learn a shared latent space, in the second stage, the Encoders and Decoders are frozen, and the Translator captures inter-variable interactions for prediction. Besides, by introducing a self-attention mechanism for multivariable fusion in the latent space, the performance achieves further improvements. Empirically, extensive experiments show the state-of-the-art performance of our method. Specifically, it reduces the MSE for near-surface air temperature and relative humidity predictions by 28.82\% and 23.39\%, respectively.
The source code is available at \url{https://github.com/ShremG/Met2Net}.

\end{abstract}
    
\begin{NoHyper}
\def\thefootnote{*}\footnotetext{Equal contribution.~~~~~~ $^\dagger$Corresonding authors.}
\end{NoHyper}
\section{Introduction}
\label{sec:intro}

Global climate change and the increasing frequency of extreme weather events make accurate weather prediction crucial~\cite{douris2021atlas}. While traditional statistical and physical models are effective in some areas, they often struggle with complex nonlinear relationships and high-dimensional data. Recently, advancements in artificial intelligence have driven the rapid development of spatiotemporal prediction models, significantly improving weather forecasting accuracy~\cite{tan2023temporal,lin2022conditional,tan2023openstl, ling2024spacetime,guen2020disentangling,han2023precipitation}. The success of these methods is attributed not only to their focusing on temporal non-stationarity but also to their incorporation of spatial dependency to capture more complex weather patterns.

\begin{figure}
  \centering
  \vspace{-0.25em}
  \begin{subfigure}{0.49\linewidth}
    \includegraphics[width=1\linewidth]{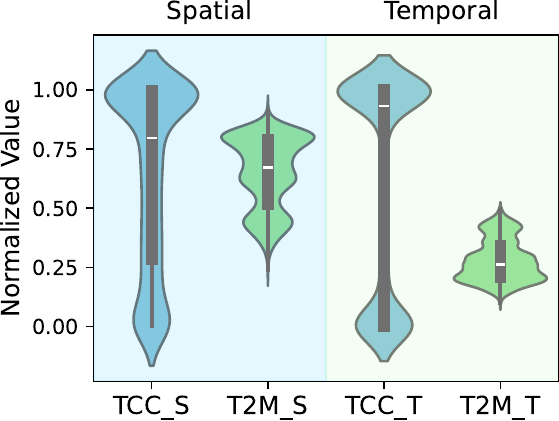} 
    \caption{\quad Data distributions.}
    \label{fig:data-a}
  \end{subfigure}
  \begin{subfigure}{0.49\linewidth}
    \includegraphics[width=1\linewidth]{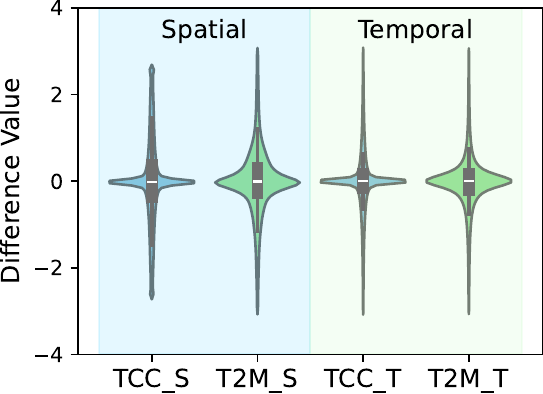} 
    \caption{\quad First-order difference.}
    \label{fig:data-b}
  \end{subfigure}
  \hfill
  \begin{subfigure}{0.49\linewidth}
     \includegraphics[width=1\linewidth,  trim=00 00 6 00, clip]{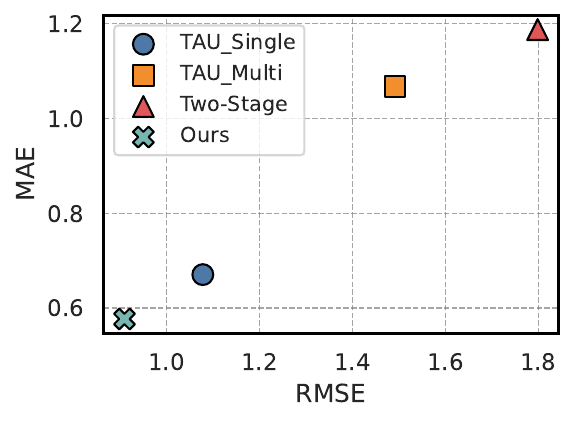} 
    \caption{\quad T2M variable.}
    \label{fig:met-a}
  \end{subfigure}
  \hfill
  \begin{subfigure}{0.49\linewidth}
     \includegraphics[width=1\linewidth,  trim=08 00 5 00, clip]{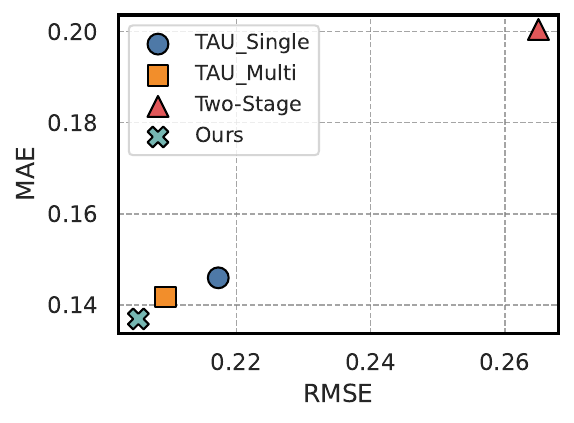} 
    \caption{\quad TCC variable.}
    \label{fig:met-b}
  \end{subfigure}
  \vspace{-0.75em}
  \caption{(a) Spatiotemporal distributions of the meteorological variables TCC and T2M highlight their heterogeneity. (b) First-order differences illustrate variations in trends and rates. (c) and (d) show that incorporating multiple variables did not enhance and even degraded TAU’s performance, while the two-stage training scheme also produced suboptimal results.}
  \label{fig:mm}
  \vspace{-1.25em}
\end{figure}
However, variables such as temperature, humidity, wind speed, and precipitation interact with each other, forming a complex climate system~\cite{bi2023accurate,staa,chen2023fuxi,chen2023fengwu,eusebi2024realistic}. The prevailing methods aim to analyze a single variable, fail to capture these dynamic relationships, thus leading to prediction errors. 
Therefore, integrating multiple meteorological variables enables the model to better understand weather patterns, identify potential extreme weather events, and provide more reliable warnings and decision support in response to climate change~\cite{staa,eusebi2024realistic}. 
\begin{figure}
  \centering
  \begin{subfigure}{0.19\linewidth}
     \includegraphics[width=1\linewidth]{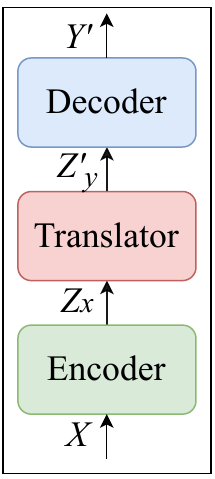} 
    \caption{E2E.}
    \label{fig:motivation-a}
  \end{subfigure}
  \begin{subfigure}{0.38\linewidth}
     \includegraphics[width=1\linewidth]{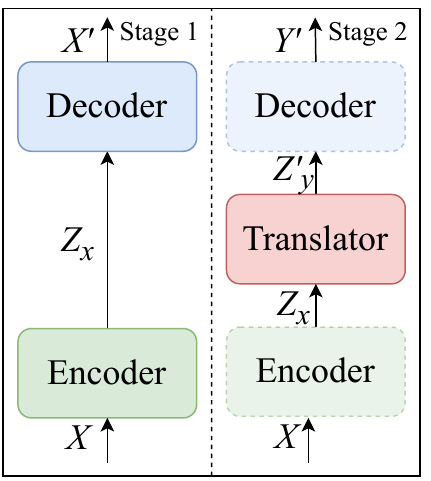} 
    \caption{Generative.}
    \label{fig:motivation-b}
  \end{subfigure}
  \begin{subfigure}{0.38\linewidth}
     \includegraphics[width=1\linewidth]{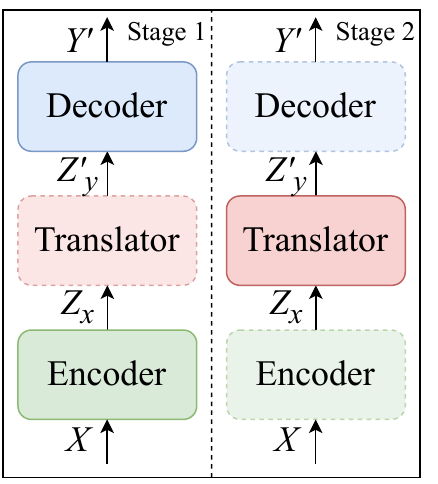} 
    \caption{Ours.}
    \label{fig:motivation-c}
  \end{subfigure}
  \vspace{-0.7em}
  \caption{(a) End-to-End (E2E) training strategy. (b) Generative model training strategy, where \textbf{\textit{task inconsistency}} exists between the two stages. (c) Met$^2$Net training strategy, which ensures task consistency across both stages. The Translator module is responsible for learning spatiotemporal features. The Translator module is responsible for learning spatiotemporal features.}
  \label{fig:motivation}
  \vspace{-10pt}
\end{figure}

During implementation, we found that simply incorporating multiple meteorological variables into the existing spatiotemporal prediction framework did not significantly improve accuracy; in some cases, it even decreased. As shown in Figure~\ref{fig:met-a} and~\ref{fig:met-b}, the experimental results confirm this performance decline. Different from image data with RGB channel, which could be regarded as multi-variables, the meteorological variables demonstrate a high degree of divergence~\cite{huang2021relative,taye2024hydrologic}. 
 As an illustration, we present the distribution of each variable's values across spatial and temporal dimensions, as shown in Figure~\ref{fig:data-a},~\ref{fig:data-b}. (Detailed experimental setting is given in Appendix.) 
These distribution divergences make it challenging for simple end-to-end training to effectively integrate these variables~\cite{nketiah2023recurrent,chen2023prompt,maina2020sensitivity}, leading to \textbf{\textit{(i) representation inconsistency}}. The reason is that when a single model processes highly divergent data, it can easily lead to a loss of the original characteristics of the data as features across different dimensions are forcibly integrated into the same space, resulting in an inconsistency between the representation and the input variables.

To address \textbf{\textit{representation inconsistency}}, two-stage training approaches, inspired by multi-modal generation models \cite{rombach2022high,esser2021taming,wu2024ivideogpt}, treat different meteorological variables as distinct modalities. This strategy, widely used in modern generative models like Latent Diffusion \cite{rombach2022high} and auto-regressive for image~\cite{yu2023magvit, cvpr2025mergevq} and video~\cite{yan2021videogpt,kondratyuk2023videopoet}, maps diverse modalities into a shared latent space, transforming heterogeneity into homogeneity \cite{qiao2024interpretable} and enhancing model performance. However, its application to spatiotemporal prediction remains rare.

When we adapted this two-stage strategy, results were unsatisfactory (Figure~\ref{fig:met-a}, \ref{fig:met-b}), which we attribute to \textbf{\textit{(ii) task inconformity}}. In generative tasks, the first stage typically employs Auto-encoders or Variational Auto-encoders  \cite{li2023comprehensive,chen2023mixed} to reconstruct data, mapping it into a latent space. The second stage then uses models like diffusion to generate representations from noise \cite{ma2024latte,yu2023video,ni2023conditional}, continuing the reconstruction process. However, in spatiotemporal prediction, while the first stage aligns with generative training, the second stage shifts to a prediction task rather than denoising reconstruction. This mismatch hinders the effectiveness of learned representations, leading to suboptimal accuracy, as shown in Figure~\ref{fig:met-a} and \ref{fig:met-b}.

To address this issue, we propose Met$^2$Net, a decoupled \textbf{two}-stage spatio-temporal forecasting model for complex \textbf{met}eorological systems, as shown in Figure~\ref{fig:motivation-c}. 
Specifically, to eliminate \textbf{\textit{representation inconsistency}} and \textbf{\textit{task inconformity}}, we freeze the gradients of the Translator in the first stage and those of the Encoder and Decoder in the second stage, ensuring alignment between training objectives. Momentum updates \cite{he2020momentum} adjust the frozen components in both stages, maintaining stability and enabling smooth parameter adaptation. Additionally, a prediction loss in the latent space ensures a unified learning objective, keeping both stages focused on forecasting rather than shifting from reconstruction to prediction. This design allows Met$^2$Net to function within a single continuous training cycle, eliminating explicit stage separation. To further improve performance, dedicated encoders and decoders preserve variable-specific features, while a self-attention mechanism in the Translator enables effective multivariable integration. By addressing both \textbf{\textit{representation inconsistency}} and  \textbf{\textit{task inconformity}}, Met$^2$Net achieves more accurate and stable multivariate spatiotemporal prediction.
To summarize, our contribution in this work is as follows:
\begin{itemize}[leftmargin=1.0em]
    \item We propose an implicit two-stage training paradigm that freezes module gradients at specific stages and employs momentum updates. By introducing prediction loss in the latent space, the approach effectively aligns training objectives, enhancing spatiotemporal prediction.
    
    \item We treat each meteorological variable as an independent modality, using separate encoders and decoders, and introduce a self-attention mechanism in the Translator to capture inter-variable relationships.

    \item We construct a general multivariate spatiotemporal prediction dataset to evaluate the robustness and generalization of our method. Empirical results show that Met$^2$Net achieves SOTA performance, reducing the MSE of T2M and R by 28.82\% and 23.39\%, resectively. Additionally, it improves hurricane trajectory prediction accuracy, demonstrating its effectiveness in weather forecasting.
\end{itemize}

\section{Related works}\label{sec:rw}
\paragraph{Spatiotemporal predictive learning.}
The advancements in recurrent-based models have greatly enhanced our understanding of spatiotemporal predictive learning.
The ConvLSTM model~\cite{shi2015convolutional} combines convolutional networks with LSTM to capture spatiotemporal patterns. Following this, PredRNN~\cite{wang2017predrnn} and PredRNN++~\cite{wang2018predrnn++} used spatiotemporal LSTM (ST-LSTM) and gradient highway units to better capture temporal dependencies and reduce gradient vanishing. MIM~\cite{wang2019memory} uses the difference between hidden states to improve handling of nonstationarity. E3D-LSTM~\cite{wang2018eidetic} adds 3D convolutions to LSTM to improve feature extraction across time and space. PhyDNet~\cite{guen2020disentangling} separates partial differential equation (PDE) dynamics from unknown factors using a recurrent physical unit. MAU~\cite{chang2021mau} introduces a motion-aware unit to capture motion-related information. Lastly, PredRNNv2~\cite{wang2022predrnn} uses a curriculum learning strategy and memory decoupling loss to improve performance.
Unlike recurrent-based methods, which have high computational costs in spatiotemporal prediction, SimVP~\cite{gao2022simvp} introduces a non-recurrent spatiotemporal prediction framework that reduces computational costs while achieving competitive performance.
Subsequently, non-recurrent models such as TAT~\cite{nie2024triplet}, TAU~\cite{tan2023temporal}, DMVFN~\cite{hu2023dynamic}, and Wast~\cite{nie2024wavelet} made further progress by incorporating triplet attention, temporal attention, dynamic multiscale voxel flow, and a 3D wavelet framework.
However, these methods are based on single-variable spatiotemporal training, and although they have achieved top-tier performance in weather prediction, they fail to fully exploit the potential of multivariable data in weather forecasting.  To address this, we propose a solution that independently encodes and decodes each variable and introduce an implicit two-stage training strategy.

\paragraph{Weather forecasting.}
Weather forecasting, as a specific application of spatiotemporal prediction, has also benefited from these advancements, leading to significant improvements in prediction accuracy and computational efficiency. Han et al.~\cite{han2023precipitation} applied the SimVP model for radar extrapolation and achieved competitive performance in precipitation prediction scenarios. Chen et al.~\cite{staa} introduced the TAU model, integrating satellite data to alleviate the issue of severe abrupt changes in precipitation scenarios. Eusebi et al.~\cite{eusebi2024realistic} utilized PINN to fuse barometric pressure for hurricane reconstruction. Jiang et al.~\cite{jiang2024applicability} employed GNN to integrate multiple variables, further enhancing the accuracy of wind speed predictions. Models such as Pangu~\cite{bi2023accurate}, FuXi~\cite{chen2023fuxi}, and Fengwu~\cite{chen2023fengwu} incorporate a wide range of meteorological variables, achieving near-perfect predictions of typhoon paths and highlighting their capability for medium-range forecasting.
This underscores the necessity and effectiveness of integrating multiple variables and multimodal data in weather forecasting.

\paragraph{Two stage training methods.}
Two-stage training is widely used in generative tasks, where models like Stable Diffusion \cite{rombach2022high}, MAGVIT \cite{yu2023magvit}, and VideoPoet \cite{kondratyuk2023videopoet} map diverse modalities into a shared latent space for improved representation learning. Typically, the first stage employs Auto-encoders (AE) or Variational Auto-encoders (VAE) \cite{li2023comprehensive,chen2023mixed} for data reconstruction, while the second stage utilizes diffusion models \cite{ma2024latte,yu2023video,ni2023conditional} to generate latent representations from noise.
Recent research on two-stage training focuses on enhancing the generality of representations learned in the first stage through large-scale pretraining~\cite{he2020momentum,grill2020bootstrap,caron2020unsupervised,oquab2023dinov2}. While this improves transferability across tasks, it requires extensive data, limiting its feasibility in scenarios with constrained datasets. 
However, in spatiotemporal prediction, the challenge lies not only in obtaining general representations but also in addressing \textit{task inconformity}, where the second stage shifts from reconstruction to forecasting. This misalignment reduces the effectiveness of learned representations for prediction, leading to suboptimal accuracy. Our work tackles this issue by designing a training strategy that better aligns the objectives of both stages, improving prediction performance without requiring large-scale pretraining.

\section{Method}
\begin{figure*}[ht]
    \vspace{-0.5em}
    \centering
    \includegraphics[width=0.99\textwidth]{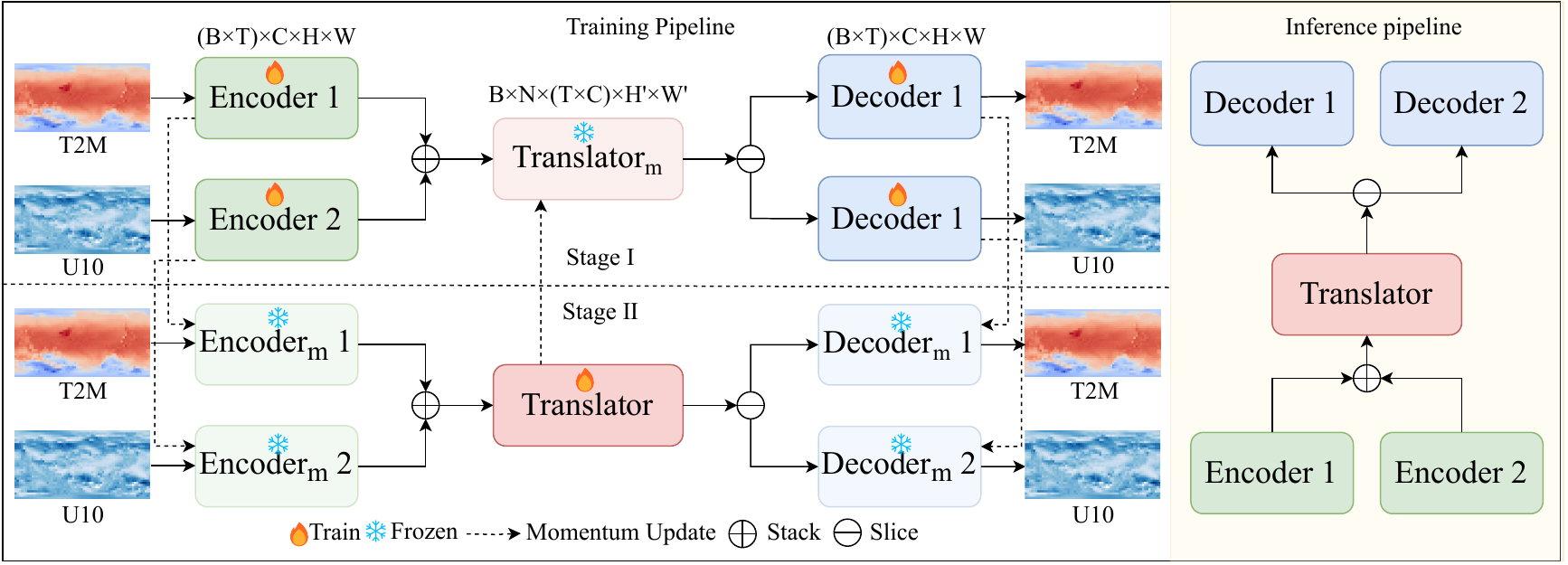} 
    \vspace{-0.5em}
    \caption{Training pipeline and Inference pipeline. The training pipeline consists of two stages: In Stage I, the Translator (blue snowflake icon) is frozen while the Encoder and Decoder (orange flame icon) are trained. In Stage II, the Encoder and Decoder are frozen, and the Translator is trained. Momentum updates, represented by dashed arrows, are applied to the frozen parameters at the end of each stage. Multiple encoders and decoders are used for different meteorological variables, enabling independent feature extraction and reconstruction. The inference pipeline uses the trained Encoder, Translator, and Decoder to generate the final output.}
    \label{fig:workflow}
    \vspace{-10pt}
\end{figure*}
\subsection{Preliminaries}
We formally define the multivariable spatiotemporal predictive learning problem as follows.
Given a multivariable time sequence \( X_{t-T+1}^t = \{ x_i^t \}_{t-T+1}^{t} \) at time \( t \) with the past \( T \) frames, we aim to predict the subsequent \( T' \) frames \( Y_{t+1}^{t+T'} = \{ y_i^{t+1} \}_{t+1}^{t+T'} \) from time \( t + 1 \), represents a collection of \( N \) variables, each with channels \( C \), height \( H \), and width \( W \). In practice, we represent the time sequences as tensors, i.e., \( X_{t-T+1}^t \in \mathbb{R}^{T \times N \times C \times H \times W} \) and \( Y_{t+1}^{t+T'} \in \mathbb{R}^{T' \times N \times C \times H \times W} \). For example, the wind variable has two channels, U and V, and its dimensions can be represented as $\mathbb{R}^{T \times 1 \times 2 \times H \times W}$.

In this paper, we denote the Encoder by \( \mathcal{E} \), the Decoder by \( \mathcal{D} \), and the Translator by \( \mathcal{T} \).
We denote the true input data by \( X \) and the label by \( Y \). The reconstructed or predicted corresponding data are represented by \( X' \) and \( Y' \), respectively. \( Z_x \) represents the encoded state of \( X \) processed by \( \mathcal{E} \), and \( Z_y \) represents the encoded state of \( Y \) processed by \( \mathcal{E} \). \( Z_y' \) denotes the predicted state of \( Z_x \) processed by \( \mathcal{T} \).


\subsection{Overview}
As shown in Figure~\ref{fig:workflow}, training is divided into two stages. In the first stage, the Translator is frozen while the Encoder and Decoder are trained to focus on spatial feature compression and reconstruction. In the second stage, the Encoder and Decoder are frozen, and a latent space prediction loss is introduced to train the Translator for inter-variable relationship modeling and prediction. During inference, only the fully trained modules are used, excluding those with frozen gradients. Thus, while this approach adds parameters and computation during training, the parameter count and computational load during inference remain the same as a standard end-to-end model. To better handle multi-variable scenarios, each variable has its own Encoder-Decoder pair.

\subsection{Implicit two-stage training strategy}\label{m1} 
Inspired by the two-stage training approach used in multimodal models in the generative domain~\cite{rombach2022high,kondratyuk2023videopoet,yu2023magvit}, we propose an implicit two-stage training strategy. This strategy involves freezing the gradients of different components at various stages within a single training process and using momentum updates to adjust parameters. While ensuring consistency of objectives between the two stages, this method retains the advantages of traditional two-stage training strategies, such as effective module transferability and efficient modality fusion.

In implementation, we should stop gradients of Translator ($\mathcal{T}$) during the first stage and gradients of Encoder ($\mathcal{E}$) \& Decoder ($\mathcal{D}$) during the second stage. However, the Translator and Encoder \& Decoder cannot be optimized alternatively with naive stop-gradient operation. 
To address this issue, we introduce the momentum update method to update the corresponding parameters. Formally, denoting the parameters of \(f_m\) and \(f\) as \(\theta_m\) and \(\theta\), we update \(\theta_m\) by:
\begin{equation}
\theta_m \leftarrow \alpha\theta_m + (1 - \alpha)\theta
\end{equation}
Here \(\alpha \in [0, 1]\) is a momentum coefficient. \(f\) represents the components \(\mathcal{E}\), \(\mathcal{D}\), and \(\mathcal{T}\) whose parameters are updated through backpropagation, while \(f_m\) represents the corresponding components with frozen gradients. 

During the training process, the objective function for the first stage is consistent with the end-to-end model, aiming to minimize the gap between $Y^{\prime}$ and $Y$. The objective for the second stage is to minimize the gap between $Z_y^{\prime}$ and $Z_y$. Ultimately, our loss function can be expressed as:
\begin{equation}
\mathcal{L} = \mathcal{L}_1(Y^{\prime},Y) + \mathcal{L}_2(Z_y^{\prime},Z_y)
\end{equation}
where $\mathcal{L}_1$ and $\mathcal{L}_2$ represent the loss functions for the first and second stages, respectively, and both are defined as Mean Squared Error (MSE) in this study. $Y$ represents the target data, $Y^{\prime}$ is the model's output data, $Z_y^{\prime}$ is the output of the second stage $\mathcal{T}$, and $Z_y$ is the output of $Y$ after being processed by $\mathcal{E}$.

\begin{table*}[th]
\centering
\caption{Quantitative comparison of predictions across multiple meteorological variables, including UV10, T2M, TCC, and R, measured using MSE, MAE, and RMSE metrics. \# Params denote the parameters of the model during the inference stage. N represents the number of models required to predict the four variables. $\downarrow$ indicates lower is better.}
\vspace{-0.5em}
\setlength{\tabcolsep}{1.5mm}
\resizebox{0.99\linewidth}{!}{
\begin{tabular}{ll|cc|ccc|ccc|ccc|ccc}
\toprule
\multicolumn{1}{c}{Method} &         Date& \#Params. & N & \multicolumn{3}{c|}{UV10}                             & \multicolumn{3}{c|}{T2M}                            & \multicolumn{3}{c|}{TCC}                            & \multicolumn{3}{c}{R}                                \\
                                     && (M)       &   & MSE$\downarrow$             & MAE$\downarrow$              & RMSE$\downarrow$             & MSE$\downarrow$              & MAE$\downarrow$              & RMSE$\downarrow$             & MSE$\downarrow$              & MAE$\downarrow$              & RMSE$\downarrow$             & MSE$\downarrow$               & MAE$\downarrow$              & RMSE$\downarrow$             \\ \hline
ConvLSTM\cite{shi2015convolutional}  &NeurIPS'2015& 14.98     & 4 & 1.8976          & 0.9215          & 1.3775          & 1.5210          & 0.7949          & 1.2330          & 0.0494          & 0.1542          & 0.2223          & 35.1460          & 4.0120          & 5.9280          \\
 PredRNN\cite{wang2017predrnn}& NeurIPS'2017& 23.57& 4& 1.8810& 0.9068& 1.3715& 1.3310& 0.7246& 1.1540& 0.0550& 0.1588& 0.2346& 37.6110& 4.0960&6.1330\\
PredRNN++\cite{wang2018predrnn++}    &ICML'2018& 38.40     & 4 & 1.8727          & 0.9019          & 1.3685          & 1.4580          & 0.7676          & 1.2070          & 0.0548          & 0.1544          & 0.2341          & 45.9930          & 4.7310          & 6.7820          \\
 MAU\cite{chang2021mau}& NeurIPS'2021& 5.46& 4& 1.9001& 0.9194& 1.3784& 1.2510& 0.7036& 1.1190& 0.0496& 0.1516& 0.2226& 34.5290& 4.0040&5.8760\\
SimVP\cite{gao2022simvp}             &CVPR'2022& 14.67     & 4 & 1.9993          & 0.9510          & 1.4140          & 1.2380          & 0.7037          & 1.1130          & 0.0477          & 0.1503          & 0.2183          & 34.3550          & 3.9940          & 5.8610          \\
ConvNeXt\cite{liu2022convnet}&CVPR'2022& 10.09& 4 & 1.6914& 0.8698& 1.3006& 1.2770& 0.7220& 1.1300& 0.0474& 0.1487& 0.2178& 33.1790& 3.9280& 5.7600\\
 HorNet\cite{rao2022hornet}&  NeurIPS'2022& 12.42& 4& 1.5539& 0.8254& 1.2466& 1.2010& 0.6906& 1.0960& 0.0469& 0.1475& 0.2166& 32.0810& 3.8260&5.6640\\
TAU\cite{tan2023temporal}            &CVPR'2023& 12.22     & 4 & 1.5925          & 0.8426          & 1.2619          & 1.1620          & 0.6707          & 1.0780          & 0.0472          & 0.1460          & 0.2173          & 31.8310          & 3.8180          & 5.6420          \\
 Wast\cite{nie2024wavelet}& AAAI'2024& 14.46& 4& -& -& -& 1.0980& 0.6338& 1.0440& -& 0.1452& 0.2150& -& 3.6940&5.5690\\
\textbf{Met$^2$Net} &\textbf{Ours}& 8.65     & 1 & \textbf{1.5055} & \textbf{0.8196} & \textbf{1.2270} & \textbf{0.8271} & \textbf{0.5770} & \textbf{0.9094} & \textbf{0.0422} & \textbf{0.1370} & \textbf{0.2054} & \textbf{24.3854} & \textbf{3.3209} & \textbf{4.9382} \\
\bottomrule
\end{tabular}
}
\label{table:comp1}
\vspace{-10pt}
\end{table*}

\begin{table}[ht]
\centering
\small
\caption{The statistics of datasets. The training or testing set has \(N_{\text{train}}\) or \(N_{\text{test}}\) samples, composed by \(T\) or \(T'\) images with the shape \((C, H, W)\). The weather refers to the multivariate prediction dataset. The subscript $_L$denotes low spatial resolution, $_{HA}$ represents high-altitude variables, $_H$and indicates high spatial resolution. The MvMmfnist is a dataset we constructed for the general multivariate spatiotemporal prediction scenario.}
\vspace{-0.5em}
\begin{tabular}{lccccccc}
\toprule
 & \(N_{\text{train}}\) & \(N_{\text{test}}\) &\(C\) &\(H\) &\(W\)   & \(T\) & \(T'\) \\
\midrule
Weather$_L$  & 52559 & 17495 & 5 & 32 & 64  & 12 & 12 \\
Weather$_{HA}$ & 54019  & 2883 & 12 & 32 & 64   & 4 & 4 \\
Weather$_H$ & 52559  & 17495 & 5 & 64 & 128  & 12 & 12  \\
ERA5 & 43801  & 8737 & 4 & 128 & 128  & 12 & 12  \\
UV10 & 52559  & 17495 & 1 & 32 & 64  & 12 & 12  \\
T2M & 52559  & 17495 & 1 & 32 & 64 & 12 & 12  \\
TaxiBJ  & 19627 & 1334 & 2 & 32 & 32  & 4  & 4  \\
MvMmfnist & 10000  & 10000 & 3 & 64 & 64  & 10 & 10  \\
\bottomrule
\end{tabular}
\label{table:datasets}
\vspace{-10pt}
\end{table}

\subsection{Multiple meteorological variables fusion}\label{m2}
In meteorological forecasting, integrating multiple variables is essential in that meteorological factors, as different physical quantities, are mutually coupled and influence each other within an actual geophysics dynamic system \cite{liu2023unified,ma2023histgnn,lam2023learning,jiang2024applicability,wang2021review}.
Merging different variables can be regarded as multimodal fusion, and we thus, enlighten by the latest techniques in multimodal fusion \cite{ho2022imagen,radford2021learning,kondratyuk2023videopoet,blattmann2023align}, design independent $\mathcal{E}$ and $\mathcal{D}$ for each meteorological variable and then used a $\mathcal{T}$ in the latent space to achieve the fusion and prediction.

Each  meteorological variable is assigned a pair of $\mathcal{E}$ and $\mathcal{D}$ with the same configuration. 
For $\mathcal{E}$ and $\mathcal{D}$, we use a two-dimensional architecture. 
$\mathcal{E}$ independently compresses each variable into the latent space along the spatial dimensions, while $\mathcal{D}$ restores the embeddings from the latent space back to the original space.
Formally, denoting the two input meteorological variables as $X_1$, $X_2$, the fusion steps read
\begin{align}
Z_1 &= \mathcal{E}_1(X_1), 
Z_2 = \mathcal{E}_2(X_2)  \\
Z &= \mathrm{Stack}(Z_1,Z_2) \\
\hat{Z^\prime} &= \mathcal{T}(Z) \\
\hat{X^\prime_1} &= \mathcal{D}_1(\mathrm{Slice}(\hat{Z^{\prime}})),  
\hat{X^\prime_2} = \mathcal{D}_2(\mathrm{Slice}(\hat{Z^{\prime}}))  
\end{align}
Here, $\mathrm{Stack}(\cdot)$ and $\mathrm{Slice}(\cdot)$ represent stacking and slicing operations, respectively. $\hat{X^\prime_1}$, $\hat{X^\prime_2}$ represent the model's prediction results for the two meteorological variables.

\subsection{Translator}\label{m4}
To model the complex dynamic dependencies of multiple meteorological variables, we utilize an attention mechanism to effectively capture the correlations between variables \cite{gangopadhyay2021spatiotemporal} and achieve effective decoupling between multiple variables.

Before processing with the $\mathcal{T}$, the embeddings of each variable for all time steps need to be arranged sequentially along the channel dimension.
For a spatiotemporal signal with $N$ variables, the length of $T$, and the spatial resolution of $(H, W)$, which is denoted by $Z \in \mathbb{R}^{N\times T \times H \times W}$, the attention matrix over variable axis reads as:
\begin{equation}
\mathbf{A} = \mathrm{softmax}(\frac{QK^\intercal}{\sqrt{d}}),
\end{equation}
where $Q, K \in\mathbb{R}^{N \times D \times H \times W}$ are query and key, which are extracted by two different 2D-CNNs, and $\sqrt{d}$ is a scaling term. 
After the $\mathrm{softmax}(\cdot)$ function, $\mathbf{A} \in \mathbb{R}^{N \times N}$ demonstrates the correlations between different variables. Subsequently, the fusion is performed by $Z_a = \mathbf{A}V$, where $V \in \mathbb{R}^{N \times T \times H \times W}$ is the value term, extracted by another 2D-CNN, $Z_a\in\mathbb{R}^{N \times T \times H \times W}$ is the fused embeddings. We denote the above operations using $\mathrm{VA}(\cdot)$ (Variable Attention). Next, we use a spatiotemporal feature extractor written as $\mathrm{ST}(\cdot)$ to extract spatiotemporal features and map them into predictions, as
\begin{equation}
\mathcal{T}(Z) = \mathrm{ST}(\mathrm{VA}(Z))
\end{equation}
In this paper, we utilize TAU Block\cite{tan2023temporal} as the practical implementation of $\mathrm{ST}(\cdot)$. We discuss the impact of using different blocks on the results in the appendix.

\section{Experiment}

In this section, we present experiments demonstrating the effectiveness of our method across two tasks: multivariate weather prediction and general spatiotemporal prediction. For multivariate weather prediction, we tested low- and high-resolution spatial data as well as high-altitude data, and we conducted a hurricane trajectory prediction experiment using real ERA5 data. For general spatiotemporal prediction, both single-variable and multivariate experiments were conducted to assess the method’s generalization and versatility. The datasets used are listed in Table~\ref{table:datasets}.

The datasets Weather$_L$ and Weather$_H$ use UV10 (10-meter u and v components of wind), T2M (2-meter temperature), TCC (total cloud cover), and R (relative humidity), for a total of four variables, with the two components of wind considered as a single variable. The Weather$_{HA}$ dataset uses U (u component of wind), V (v component of wind), T (temperature), and R (relative humidity), for a total of four variables. Additionally, the ERA5 dataset includes MSL (mean sea level pressure), U10, V10, and T2M, also consisting of four variables. For the ERA5 dataset, we performed spatial cropping within the range of 110$^\circ$E–142$^\circ$E and 12.25$^\circ$N–44$^\circ$N. The dataset was split into three subsets: 2017–2021 for training, 2022 for validation, and 2023 for testing.

\begin{figure*}
  \centering
  \begin{subfigure}{0.49\linewidth}
    \includegraphics[width=1\linewidth]{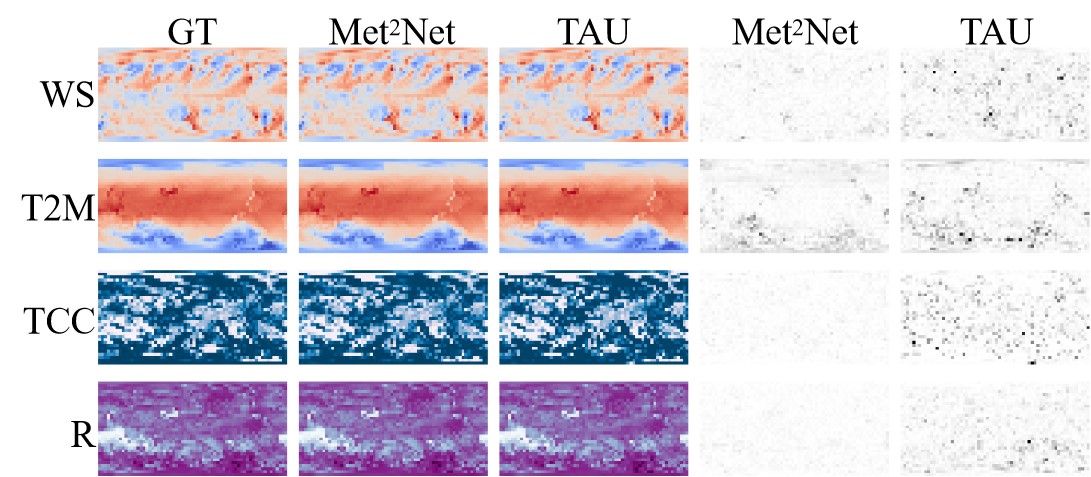} 
    \caption{t=1.}
    \label{fig:short-a}
  \end{subfigure}
  \hfill
  \begin{subfigure}{0.49\linewidth}
   \includegraphics[width=1\linewidth]{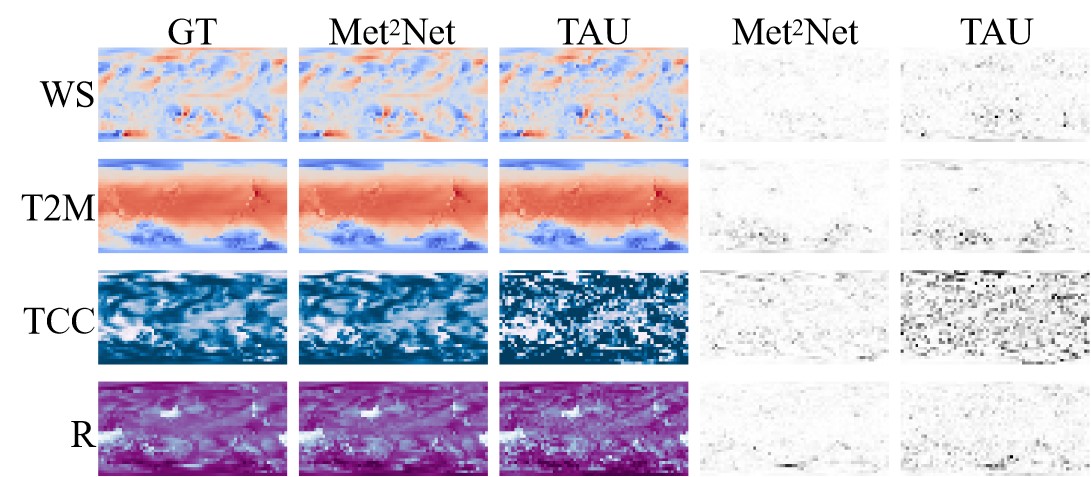} 
    \caption{t=12.}
    \label{fig:short-b}
  \end{subfigure}
  \vspace{-0.7em}
  \caption{Visualization of prediction results for different lead times. (a) Results at a forecast time of 1 hour. The background in white represents the absolute error ($| Y'-Y |$) for each model. (b) Results at a forecast time of 12 hours. Across both 1-hour and 12-hour forecasts, the error of the \textbf{TAU} model is consistently higher than that of our \textbf{Met$^2$Net} model.}
  \label{fig:case}
\end{figure*}

\begin{figure}[ht]
    \centering
    \includegraphics[width=1\linewidth]{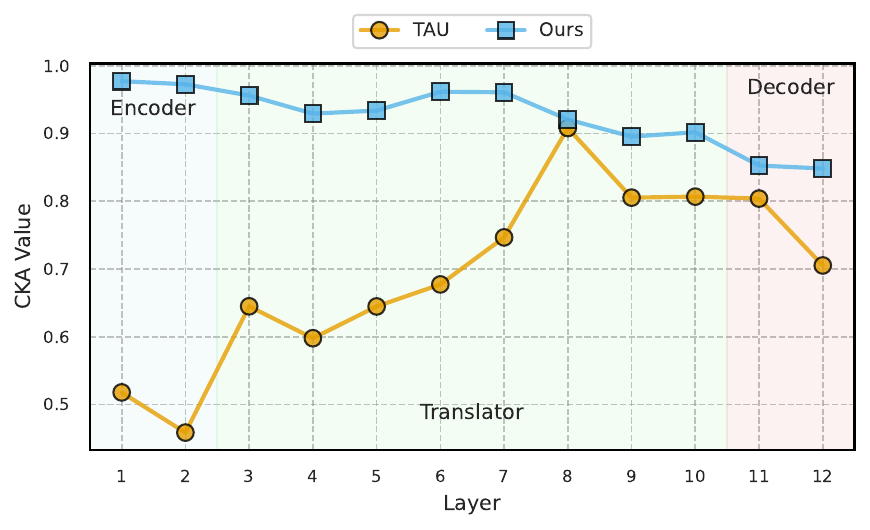} 
    \vspace{-2em}
    \caption{Met$^2$Net effectively resolves \textbf{\textit{representation inconsistency}} by maintaining higher CKA values across layers compared to TAU. The TAU model exhibits a gradual increase in CKA values through the Translator section but maintains relatively lower CKA values overall. In contrast, Met$^2$Net sustains higher CKA values across all layers, particularly in the Encoder and Translator sections, indicating improved representation consistency. }
    \label{fig:cka}
    \vspace{-1em}
\end{figure}

\begin{table*}[]
\centering
\caption{Quantitative comparison of predictions across multiple high-altitude variable meteorological variables, including R, T, U, and V, measured using MSE, MAE, and RMSE metrics. $\downarrow$ indicates lower is better.}
\vspace{-0.5em}
\setlength{\tabcolsep}{1.3mm}
\resizebox{0.95\linewidth}{!}{
\begin{tabular}{ll|ccc|ccc|ccc|ccc}
\toprule
Method                                &Date
& \multicolumn{3}{c}{R}                             & \multicolumn{3}{c|}{T}                              & \multicolumn{3}{c|}{U}                              & \multicolumn{3}{c}{V}                                \\
                                      &
& MSE$\downarrow$             & MAE$\downarrow$            & RMSE$\downarrow$            & MSE$\downarrow$             & MAE$\downarrow$             & RMSE$\downarrow$            & MSE$\downarrow$             & MAE$\downarrow$             & RMSE$\downarrow$            & MSE$\downarrow$              & MAE$\downarrow$             & RMSE$\downarrow$            \\ \hline
ConvLSTM\cite{shi2015convolutional}   &NeurIPS'2015
& 368.15          & 13.49          & 19.19           & 6.3030& 1.7695          & 2.5107          & 30.002          & 3.8923          & 5.4774          & 30.789           & 3.8238          & 5.5488          \\
 PredRNN\cite{wang2017predrnn} &NeurIPS'2017
& 354.57& 13.17& 18.830& 5.5966& 1.6411& 2.3657& 27.484& 3.6776& 5.2425& 28.973& 3.6617&5.3827\\
PredRNN++\cite{wang2018predrnn++}     &ICML'2018
& 363.15          & 13.25          & 19.056          & 5.6471          & 1.6433          & 2.3763          & 28.396          & 3.7322          & 5.3288          & 29.872           & 3.7067          & 5.4655          \\
 MAU\cite{chang2021mau} &NeurIPS'2021
& 363.36& 13.50& 19.062& 5.6287& 1.6810& 2.3725& 27.582& 3.7409& 5.2519& 27.929& 3.6700&5.2848\\
SimVP\cite{gao2022simvp}              &CVPR'2022
& 370.03          & 13.58          & 19.236          & 6.1068          & 1.7554          & 2.4712          & 28.782          & 3.8435          & 5.3649          & 29.094           & 3.7614          & 5.3939          \\
ConvNeXt\cite{liu2022convnet} &CVPR'2022
& 367.39& 13.52& 19.168& 6.1749& 1.7448& 2.4849& 29.764& 3.8688& 5.4556& 31.326& 3.8435& 5.5969\\
 HorNet\cite{rao2022hornet} &NeurIPS'2022
& 353.02& 13.02& 18.789& 5.5856& 1.6198& 2.3634& 28.192& 3.7142& 5.3096& 30.028& 3.7148&5.4798\\
TAU\cite{tan2023temporal}             &CVPR'2023& 342.63          & 12.80          & 18.510          & 4.9042          & 1.5341          & 2.2145          & 24.719          & 3.5060          & 4.9719          & 25.456           & 3.4723          & 5.0454          \\
\textbf{Met$^2$Net} &\textbf{Ours}& \textbf{337.98}& \textbf{12.74}& \textbf{18.38}& \textbf{4.1376}& \textbf{1.3688}& \textbf{2.0341}& \textbf{22.832} & \textbf{3.3449}& \textbf{4.7783}& \textbf{23.900}& \textbf{3.3597}& \textbf{4.8888}\\
\bottomrule
\end{tabular}
}
\label{table:comp2}
\end{table*}

\begin{table*}[]
\centering
\caption{Quantitative comparison of predictions across multiple high-resolution meteorological variables, including UV10, T2M, TCC, and R, measured using MSE, MAE, and RMSE metrics. $\downarrow$ indicates lower is better.}
\vspace{-0.5em}
\setlength{\tabcolsep}{1.3mm}
\resizebox{0.95\linewidth}{!}{
\begin{tabular}{ll|ccc|ccc|ccc|ccc}
\toprule
Method                                &Date
& \multicolumn{3}{c}{UV10}                             & \multicolumn{3}{c|}{T2M}                              & \multicolumn{3}{c|}{TCC}                              & \multicolumn{3}{c}{R}                                \\
                                      &
& MSE$\downarrow$             & MAE$\downarrow$            & RMSE$\downarrow$            & MSE$\downarrow$             & MAE$\downarrow$             & RMSE$\downarrow$            & MSE$\downarrow$             & MAE$\downarrow$             & RMSE$\downarrow$            & MSE$\downarrow$              & MAE$\downarrow$             & RMSE$\downarrow$            \\ \hline
ConvLSTM\cite{shi2015convolutional}   &NeurIPS'2015
& 1.5691& 0.8653& 1.2518& 1.1385& 0.6774& 1.0654& 0.0450& 0.1431& 0.2121& 28.9580& 3.6594& 5.3769\\
 PredRNN\cite{wang2017predrnn} &NeurIPS'2017
& 1.4083& 0.8037& 1.1859& 1.1862& 0.6695& 1.0873& 0.0482& 0.1444& 0.2195& 30.1590& 3.7334&5.4875\\
 MAU\cite{chang2021mau} &NeurIPS'2021
& 1.5781& 0.8578& 1.2554& 1.2493& 0.7097& 1.1161& 0.0446& 0.1451& 0.2111& 30.7881& 3.7989&5.5442\\
SimVP\cite{gao2022simvp}              &CVPR'2022
& 1.7654& 0.9332& 1.3279& 1.2750& 0.7333& 1.1278& 0.0448& 0.1451& 0.2115& 32.7849& 4.0132& 5.7217\\
ConvNeXt\cite{liu2022convnet} &CVPR'2022
& 1.4804& 0.8399& 1.2160& 1.2155& 0.7048& 1.1010& 0.0457& 0.1455& 0.2136& 31.9944& 3.8841& 5.6518\\
TAU\cite{tan2023temporal}             &CVPR'2023& 1.3038& 0.7887& 1.1411& 1.0599& 0.6592& 1.0282& 0.0434& 0.1403& 0.2082& 28.2291& 3.6307& 5.3090\\
\textbf{Met$^2$Net} &\textbf{Ours}& \textbf{1.2745}& \textbf{0.7835}& \textbf{1.1283}& \textbf{0.8936}& \textbf{0.6194}& \textbf{0.9446}& \textbf{0.0378}& \textbf{0.1270}& \textbf{0.1944}& \textbf{21.5415}& \textbf{3.1606}& \textbf{4.6382}\\
\bottomrule
\end{tabular}
}
\label{table:highres}
\end{table*}

\begin{table*}[]
\centering
\caption{Quantitative comparison of predictions performance on the ERA5 dataset, including MSL, U10, V10, and T2M, measured using MSE, MAE, and R$^2$ metrics. $\downarrow$ indicates lower is better, while $\uparrow$ indicates higher is better.}
\vspace{-0.5em}
\setlength{\tabcolsep}{1.3mm}
\resizebox{0.95\linewidth}{!}{
\begin{tabular}{ll|ccc|ccc|ccc|ccc}
\toprule
Method                                &Date
& \multicolumn{3}{c}{MSL}                             & \multicolumn{3}{c|}{U10}                              & \multicolumn{3}{c|}{V10}                              & \multicolumn{3}{c}{T2M}                                \\
                                      &
& MSE$\downarrow$             & MAE$\downarrow$            & R$^2$$\uparrow$& MSE$\downarrow$             & MAE$\downarrow$             & R$^2$$\uparrow$            & MSE$\downarrow$             & MAE$\downarrow$             & R$^2$$\uparrow$            & MSE$\downarrow$              & MAE$\downarrow$             & R$^2$$\uparrow$            \\ \hline
ConvLSTM\cite{shi2015convolutional}   &NeurIPS'2015
& 16577.42         & 90.8390         & 0.9677          & 1.7236 & 0.9238         & 0.9397        & 1.8728          & 0.9813         & 0.9060         & 2.0504          & 0.9236         & 0.9873         \\
ConvNeXt\cite{liu2022convnet} &CVPR'2022
& 15618.98& 92.7848& 0.9696& 1.5358& 0.8495& 0.9462& 1.6811& 0.9020& 0.9156& 1.2860& 0.7347& 0.9920\\
 HorNet\cite{rao2022hornet} &NeurIPS'2022
& 14585.94& 88.7772& 0.9716& 1.4964& 0.8443& 0.9476& 1.6184& 0.8896& 0.9188& 1.1962&0.7087&0.9926\\
 MogaNet\cite{li2024moganet} &ICLR'2024
& 18617.73& 100.8613& 0.9638& 1.5582& 0.8756& 0.9454& 1.6547& 0.9084& 0.9170& 1.4031& 0.7618&0.9913\\
TAU\cite{tan2023temporal}             &CVPR'2023& 16541.85          & 93.7000          & 0.9678         & 1.4979          & 0.8404          & 0.9476          & 1.6121         & 0.8824        & 0.9191          & 1.2808          & 0.7205          & 0.9920          \\
\textbf{Met$^2$Net} &\textbf{Ours}& \textbf{9645.28}& \textbf{70.3976}& \textbf{0.9812}& \textbf{1.2070}& \textbf{0.7518}& \textbf{0.9577}& \textbf{1.3124} & \textbf{0.7954}& \textbf{0.9341}& \textbf{0.8176}& \textbf{0.5866}& \textbf{0.9949}\\
\bottomrule
\end{tabular}
}
\label{table:era5}
\end{table*}

\subsection{Setup}\label{ES}
The baseline experiments were conducted using the OpenSTL \cite{tan2023openstl} library, with all meteorological variable data sourced from the WeatherBench \cite{rasp2020weatherbench} dataset. The settings followed the OpenSTL configuration, where each variable inputs 12 frames and predicts the subsequent 12 frames. The optimizer used was Adam \cite{iclr2015adam}, with a training batch size of 16, running for 50 epochs, and a momentum parameter $\alpha$ set to 0.999. We evaluated the test set using three metrics: MSE, MAE, and RMSE.

\subsection{Multiple Meteorological Variables Prediction}\label{CSOTA}

\vspace{0.5em}
\noindent\textbf{Low-Resolution.} 
Table~\ref{table:comp1} presents a quantitative comparison between our method and existing models for low-resolution multivariable meteorological prediction. Our method achieves SOTA performance across all major metrics, including MSE, MAE, and RMSE. Compared to TAU, our approach reduces MSE by 5.46\%, 28.82\%, 10.59\%, and 23.39\% for the four variables, respectively. To illustrate these differences more intuitively, Figure~\ref{fig:case} presents a visual comparison between models, where Met$^2$Net demonstrates smaller spatial prediction errors across the four variables, highlighting its advantage in accuracy.

\begin{figure}[ht]
    \centering
    \includegraphics[width=1\linewidth]{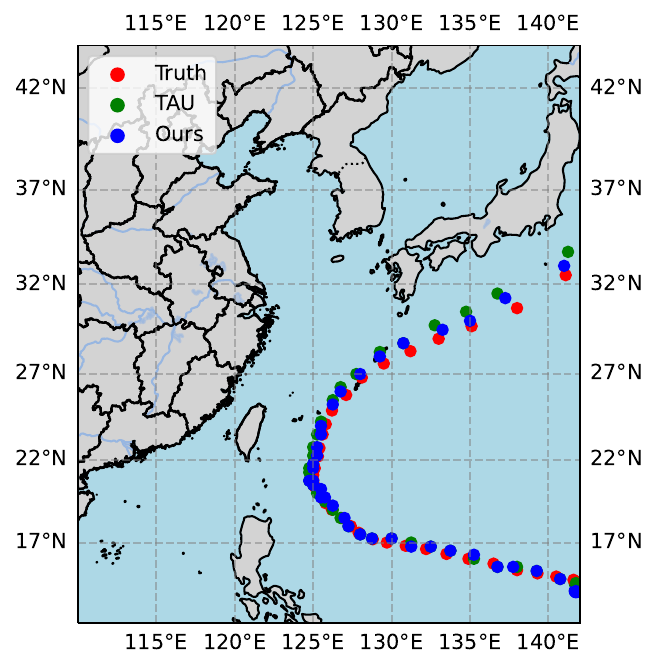} 
    \vspace{-2.5em}
    \caption{Comparison of predicted and ground truth tracks of Typhoon Mawar (3-hour lead time). The figure shows the observed trajectory alongside predictions from TAU and Met$^2$Net. Met$^2$Net more closely follows the ground truth, demonstrating improved accuracy in typhoon trajectory forecasting.}
    \label{fig:mawar}
    \vspace{-1em}
\end{figure}

\vspace{0.5em}
\noindent\textbf{High-Altitude Variable.}  
To further validate the robustness of our approach under varying atmospheric conditions, we conducted an experiment on high-altitude meteorological prediction. Table~\ref{table:comp2} compares the performance of different models in this task, showing that our method maintains SOTA performance. This experiment evaluates model performance at three altitude levels—150, 500, and 850 hPa—using 6-hour intervals. Specifically, the previous 4 frames (representing one day's data) are used to predict the next 4 frames, simulating a forecast of the following day's weather based on current data.

\vspace{0.5em}
\noindent\textbf{High-Resolution.}
To assess the model's ability to capture fine-scale meteorological patterns, we conducted an experiment on high-resolution meteorological prediction. Table~\ref{table:highres} presents a quantitative comparison across multiple high-resolution variables—UV10, T2M, TCC, and R—using MSE, MAE, and RMSE as evaluation metrics. Met$^2$Net demonstrates competitive performance across all metrics, achieving the lowest error values, which underscores its robustness and accuracy in high-resolution prediction tasks.

\vspace{0.5em}
\noindent\textbf{Tracking Tropical Cyclones.}
Our method was evaluated on Typhoon Mawar’s trajectory prediction using the ERA5 dataset, which includes MSL, U10, V10, and T2M variables. As shown in Figure~\ref{fig:mawar}, our model closely follows the ground truth while outperforming the TAU model, particularly in the early and mid-stages. Notably, it achieves higher accuracy with a 3-hour lead time while using fewer parameters. The quantitative results in Table~\ref{table:era5} further confirm its superiority, with the lowest MSE and MAE and the highest R$^2$ across all variables, demonstrating its efficiency and effectiveness in tropical cyclone forecasting.


\vspace{0.5em}
\noindent\textbf{Further analysis.}
To demonstrate the effectiveness of our method in addressing \textbf{representation inconsistency}, we analyzed the Centered Kernel Alignment (CKA)~\cite{kornblith2019similarity} values across layers for two meteorological variables: T2M and TCC. As shown in Figure~\ref{fig:cka}, this analysis compares CKA values for both variables in the TAU and Ours models, focusing on the Encoder, Translator, and Decoder sections. CKA values, which measure similarity between layer representations, reflect each model’s alignment with the internal structure of T2M and TCC, where higher values indicate stronger representation consistency. The TAU model shows a gradual increase in CKA values through the Translator section, whereas the Ours model maintains consistently higher values, particularly in the Encoder and Translator sections. This suggests that the Ours model better preserves representation consistency across variables.

\begin{table}[ht]
\centering
\caption{Quantitative comparison on the MvMmfnist. The learning rate was optimized by selecting the best value from $\{1e^{-4}$, $1e^{-3}$, $5e^{-3}\}$.}
\vspace{-0.5em}
\begin{tabular}{lcccl}
\hline
Method& MSE$\downarrow$& MAE$\downarrow$& SSIM$\uparrow$&PSNR$\uparrow$\\
\hline
SimVP      & 155.52& 397.10& 0.8812&19.2825\\
ConvNeXt    & 174.97& 434.63& 0.8682&18.7304\\
TAU        & 147.69& 385.13& 0.8833&19.5075\\
\textbf{Met$^2$Net} & \textbf{132.89}& \textbf{328.67}& \textbf{0.9086}& \textbf{19.9069}\\
\hline
\end{tabular}
\label{table:mvmnist}
\end{table}

\subsection{General Spatiotemporal Prediction}\label{gsp}
To evaluate the generalization capability of our model, we conducted experiments on single-variable and multivariable spatiotemporal prediction datasets. TaxiBJ was chosen for single-variable prediction, representing urban traffic flow, while a three-channel MvMmfnist dataset was created for multivariable prediction. These experiments aim to assess the model's ability to handle diverse data types and spatiotemporal patterns, validating its applicability across different domains.

\vspace{0.5em}
\noindent\textbf{MvMmfnist.} 
The dataset simulates the complexity of multivariate spatiotemporal prediction scenarios. It contains three channels, each functioning as an independent moving dataset. To better represent real-world scenarios, such as multivariate weather prediction, the channels are interrelated yet retain some independence. Specifically, all channels share the same movement trajectory: the second channel’s moving object is from the Fashion MNIST~\cite{xiao2017fashion} dataset instead of MNIST~\cite{srivastava2015unsupervised}, while the third channel is generated by flipping the foreground and background of the first channel. This design introduces both unique features and interrelationships among the channels.
Experiments on this dataset, with results shown in Table~\ref{table:mvmnist}, indicate that the Met$^2$Net achieved state-of-the-art performance, demonstrating its effectiveness in multivariate spatiotemporal prediction scenarios. All models were trained using the Adam optimizer for 200 epochs, with the best-performing epoch selected for evaluation, ensuring a fair comparison.

\vspace{0.5em}
\noindent\textbf{TaxiBJ.} Table~\ref{table:taxibj} shows experimental results on the TaxiBJ dataset, where the only difference between our method, Met$^2$Net, and the TAU model is the training strategy, with all other configurations kept identical. Results indicate that Met$^2$Net achieves comparable performance to TAU across all metrics, including MSE, MAE, SSIM, and PSNR. This similarity suggests that our training approach provides advantages in multivariable prediction tasks without compromising performance in single-variable scenarios.
\begin{table}[ht]
\centering
\small
\caption{Quantitative comparison on the Taxibj. The subscript $_S$ in Baseline$_S$ indicates the single-variable model.}\vspace{-0.5em}
\begin{tabular}{lcccl}
\hline
Method& MSE$\downarrow$& MAE$\downarrow$& SSIM$\uparrow$&PSNR$\uparrow$\\
\hline
ConvLSTM\cite{shi2015convolutional}    & 0.3358& 15.32& 0.9836&39.45\\
PredRNN\cite{wang2017predrnn}& 0.3194& 15.31& 0.9838&39.51\\
PredRNN++\cite{wang2018predrnn++}   & 0.3348& 15.37& 0.9834&39.47\\
MAU\cite{chang2021mau} & 0.3268& 15.26& 0.9834&39.52\\
SimVP\cite{gao2022simvp}       & 0.3282& 15.45& 0.9835&39.45\\
MogaNet\cite{li2023moganet}     & 0.3114& 15.06& 0.9847&39.70\\
HorNet\cite{rao2022hornet}& 0.3186& 15.01& 0.9843&39.66\\
TAU\cite{tan2023temporal}         & 0.3108& 14.93& 0.9848&39.74\\
\textbf{Met$^2$Net$_S$} & 0.3164& 14.82& 0.9851&39.81\\
\hline
\end{tabular}
\label{table:taxibj}
\end{table}

\begin{table}[ht]
\centering
\caption{Ablation study of the proposed method on MSE. MED (Multiple Encoder and Decoder). ITS (Implicit two-stage). The subscript $_S$ in Baseline$_S$ indicates the single-variable model.} \vspace{-0.5em}
\begin{tabular}{lcccc}
\hline
Method& UV10& T2M& TCC& R\\
\hline
Baseline$_S$ & 1.5925& 1.1620& 0.0472&31.8310\\
Baseline    & 1.8347 & 2.2247 & 0.0439 & 29.0749 \\
+MED        & 1.8327 & 1.6708 & 0.0445 & 28.8003 \\
+VA(\(\cdot\)) & 1.6287 & 1.5432 & 0.0428 & 26.6849 \\
+ITS        & \textbf{1.5055} & \textbf{0.8247} & \textbf{0.0422} & \textbf{24.3854} \\
\hline
\end{tabular}
\label{table:ablation} \vspace{-1em}
\end{table}
\subsection{Ablation Study}\label{as}
Table~\ref{table:ablation} presents ablation study results for the proposed method in multivariable prediction. Using the TAU model as a baseline for multivariate meteorological forecasting, results indicate that model performance declines when handling multiple variables compared to single-variable predictions. This suggests that complex variable interactions challenge model precision, highlighting the need for optimized multivariable fusion strategies.

Starting from the baseline model, we progressively introduced Multi-Encoder-Decoder (MED), Variable Attention (VA), and Implicit Two-Stage Training (ITS). Results show a consistent MSE decrease across all variables, with substantial improvements after incorporating MED and VA, indicating that these enhancements decouple variable relationships and significantly improve accuracy. Further gains with ITS confirm the effectiveness of our training strategy.
\section{Conclusion}\label{conclusion}
In this paper, we proposed a novel implicit two-stage training paradigm to improve spatiotemporal prediction tasks involving multiple meteorological variables. By treating each variable as an independent modality and incorporating separate encoders, decoders, and a self-attention mechanism, our framework, Met$^2$Net, effectively captures complex interactions between variables and enhances prediction accuracy. We also constructed a general multivariate spatiotemporal prediction dataset, demonstrating the robustness and generalization of our method across different tasks.


\section*{Acknowledgement}
This research was partly supported by the Sichuan Science and Technology Achievement Transfer and Transformation Demonstration Project (2024ZHCG0026), the Sichuan Science and Technology Program (2024NSFJQ0035), and the Talents Program supported by the Organization Department of the Sichuan Provincial Party Committee.

{
    \small
    \bibliographystyle{ieeenat_fullname}
    \bibliography{main}
}
\clearpage
\setcounter{page}{1}
\maketitlesupplementary

\section{Appendix}
\label{sec:Appendix}

\begin{algorithm}[t]
\caption{Pseudocode of Implicit Two-Stage Process in a PyTorch-like Style Integrated Within a Inference Pipeline.}
\label{alg:test}
\definecolor{codeblue}{rgb}{0.25,0.5,0.5}
\lstset{
  backgroundcolor=\color{white},
  basicstyle=\fontsize{7.2pt}{7.2pt}\ttfamily\selectfont,
  columns=fullflexible,
  breaklines=true,
  captionpos=b,
  commentstyle=\fontsize{7.2pt}{7.2pt}\color{codeblue},
  keywordstyle=\fontsize{7.2pt}{7.2pt},
}
\begin{lstlisting}[language=python]
# E1, E2: encoders.
# D1, D2: decoders.
# H: translator.

for x in loader:  # load a minibatch x with N samples
    x1, x2 = slice(x) # Slicing Operations in Python
    z1_x = E1(x1) # Independent encoding of Variables
    z2_x = E2(x2)
    z_x = torch.stack(z1_x,z2_x)
    
    # Spatio-Temporal learnning and variable fusion
    z_y = H(z_x) 
    
    z1_y, z2_y = slice(z_y)
    y1 = D1(z1_y) # Independent decoding of Variables
    y2 = D2(z2_y)
    y_pre =  torch.stack(y1, y2)
    
\end{lstlisting}
\end{algorithm}

\begin{algorithm}[t]
\caption{Pseudocode of Implicit Two-Stage Process in a PyTorch-like Style Integrated Within a Training Pipeline.}
\label{alg:train}
\definecolor{codeblue}{rgb}{0.25,0.5,0.5}
\lstset{
  backgroundcolor=\color{white},
  basicstyle=\fontsize{7.2pt}{7.2pt}\ttfamily\selectfont,
  columns=fullflexible,
  breaklines=true,
  captionpos=b,
  commentstyle=\fontsize{7.2pt}{7.2pt}\color{codeblue},
  keywordstyle=\fontsize{7.2pt}{7.2pt},
}
\begin{lstlisting}[language=python]
# E1, E2, E1_m, E2_m: encoder that applies gradient updates and momentum updates to two different variables.
# D1, D2, D1_m, D2_m: decoder that applies gradient updates and momentum updates to two different variables.
# H, H_m: translator that gradient updates and momentum updates.
# a: momentum

E1_m.params = E1.params  # initialize
E2_m.params = E2.params  # initialize
D1_m.params = D1.params  # initialize
D2_m.params = D2.params  # initialize
H_m.params = H.params  # initialize

for x,y in loader:  # load a minibatch x,y with N samples
    # stage 1
    x1, x2 = slice(x) # Slicing Operations in Python
    z1_x = E1(x1) # Independent encoding of Variables
    z2_x = E2(x2)
    z_x = torch.stack(z1_x,z2_x)
    
    # Spatio-Temporal learnning and variable fusion
    z_y = H_m(z_x) 
    
    z1_y, z2_y = slice(z_y)
    y1 = D1(z1_y) # Independent decoding of Variables
    y2 = D2(z2_y)
    y_rec =  torch.stack(y1, y2)
    
    # momentum update
    E1_m.params = a*E1_m.params +(1-a)*E1.params  
    E2_m.params = a*E2_m.params +(1-a)*E2.params 
    D1_m.params = a*D1_m.params +(1-a)*D1.params  
    D2_m.params = a*D2_m.params +(1-a)*D2.params   

    loss_rec = MSE(y_rec,y)

    # stage 2
    x1, x2 = slice(x) # Slicing Operations in Python
    z1_x = E1_m(x1) # use momentum updates module
    z2_x = E2_m(x2)
    z_x = torch.stack(z1_x, z2_x)
    
    # Spatio-Temporal learnning and variable fusion
    z_y_pre = H(z_x) 
    
    z1_y, z2_y = slice(z_y)
    y1 = D1_m(z1_y) 
    y2 = D2_m(z2_y)
    y_pre=  torch.stack(y1, y2)

    y1, y2 = slice(y)
    z1_y = E1_m(y1) # use momentum updates module
    z2_y = E2_m(y2)
    z_y = torch.stack(z1_y, z2_y)

    # momentum update
    H_m.params = a*H_m.params +(1-a)*H.params  

    loss_pre = MSE(z_y_pre,z_y)

    loss = loss_rec + loss_pre
    
    # Adam update: query network
    loss.backward()
    
\end{lstlisting}
\end{algorithm}
\subsection{Experimental Setup for Variable Distributions}\label{app:datadiff}
\paragraph{Data Distributions.}The experiment utilized the 2018 T2M (2-meter air temperature) and TCC (Total Cloud Cover) data from the WeatherBench dataset to analyze spatial and temporal distribution patterns. The data were normalized to the range [0, 1] for comparability. Spatial analysis was based on grid data from a single time step to capture geographic variability, while temporal analysis focused on the time series data of a single grid point.
\paragraph{First-Order Differences.}The experiment analyzed first-order differences of 2018 T2M and TCC data from the WeatherBench dataset. Temporal differences were calculated between steps, and spatial differences from adjacent grid points along the h and w dimensions. Differences were standardized to zero mean and unit variance, with outliers exceeding three standard deviations removed.

\subsection{Pseudocode of Inference Pipeline}\label{app:c}
In Algorithm~\ref{alg:test}, we present the pseudocode of our method within a inference pipeline. For simplicity, we demonstrate the case with only two variables.

\subsection{Pseudocode of Trainning Pipeline}\label{app:alg1}
In Algorithm~\ref{alg:train}, we present the pseudocode of our method within a training pipeline. For simplicity, we demonstrate the case with only two variables.

\subsection{Impact of different blocks in translator}\label{app:translator}
We tested different blocks within the Translator of our framework, as shown in Table~\ref{tab:one}. The results indicate that while the TAU block remains a competitive choice, our method consistently outperforms the baseline methods across all tested blocks. This demonstrates the robustness of our framework in handling various blocks. Regardless of the block selected, our method maintains superior performance, validating the effectiveness of the proposed framework across different configurations.

\begin{table}[h]
\centering
\caption{Impact of using different \textbf{Blocks} in the translator on T2M and TCC prediction performance. The light gray background indicates results not applied in our framework. The white background indicates results obtained using different translators within our framework.}
\resizebox{\linewidth}{!}{
\begin{tabular}{l|ccc|ccc}
\toprule
Method     & \multicolumn{3}{c|}{T2M} & \multicolumn{3}{c}{TCC}  \\
           & MSE    & MAE    & RMSE   & MSE    & MAE    & RMSE   \\
\midrule
\cellcolor{gray!20} HorNet   & \cellcolor{gray!20} 1.2010 &\cellcolor{gray!20} 0.6906 &\cellcolor{gray!20} 1.0960 &\cellcolor{gray!20} 0.0469 &\cellcolor{gray!20} 0.1475 &\cellcolor{gray!20} 0.2166 \\
\cellcolor{gray!20} TAU   &\cellcolor{gray!20} 1.1620 &\cellcolor{gray!20} 0.6707 &\cellcolor{gray!20} 1.0780 &\cellcolor{gray!20} 0.0472 &\cellcolor{gray!20} 0.1460 &\cellcolor{gray!20} 0.2173 \\
\cellcolor{gray!20} Wast   &\cellcolor{gray!20} 1.0980 &\cellcolor{gray!20} 0.6338 &\cellcolor{gray!20} 1.0440 &\cellcolor{gray!20} - &\cellcolor{gray!20} 0.1452 &\cellcolor{gray!20} 0.2150 \\
\midrule
ConvNext   & 1.0238 & 0.6598 & 1.0105 & 0.0440 & 0.1426 & 0.2096 \\
SimVPv2    & 0.9215 & 0.6148 & 0.9588 & 0.0425 & \textbf{0.1367} & 0.2061 \\
PoolFormer & 0.9493 & 0.6271 & 0.9730 & 0.0435 & 0.1426 & 0.2085 \\
Hornet     & 0.8778 & 0.5987 & 0.9358 & 0.0423 & 0.1388 & 0.2055 \\
Moga       & 1.0314 & 0.6643 & 1.0141 & 0.0445 & 0.1455 & 0.2109 \\
TAU        & \textbf{0.8271} & \textbf{0.5770} & \textbf{0.9094} & \textbf{0.0422} & 0.1370 & \textbf{0.2054}\\
\bottomrule
\end{tabular}
}
\label{tab:one}
\end{table}

\subsection{Performance evolves with time steps.}\label{app:timestep}
\begin{figure}[h]
  \centering
  \begin{subfigure}{0.49\linewidth}
    \includegraphics[width=1\linewidth]{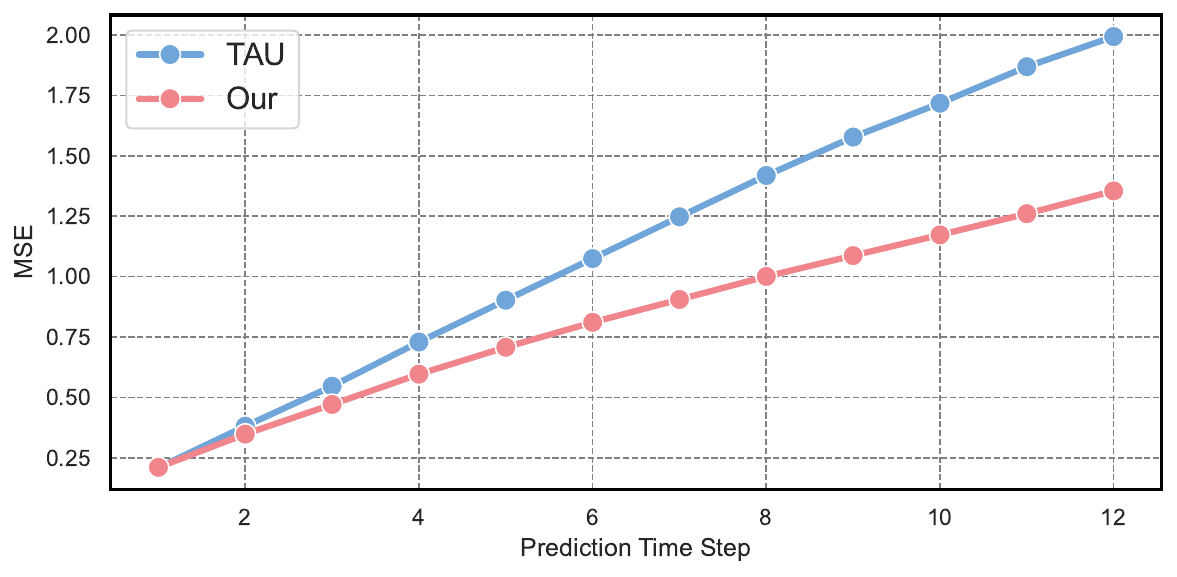} 
    \caption{T2M on MSE.}
  \end{subfigure}
  \begin{subfigure}{0.49\linewidth}
    \includegraphics[width=1\linewidth]{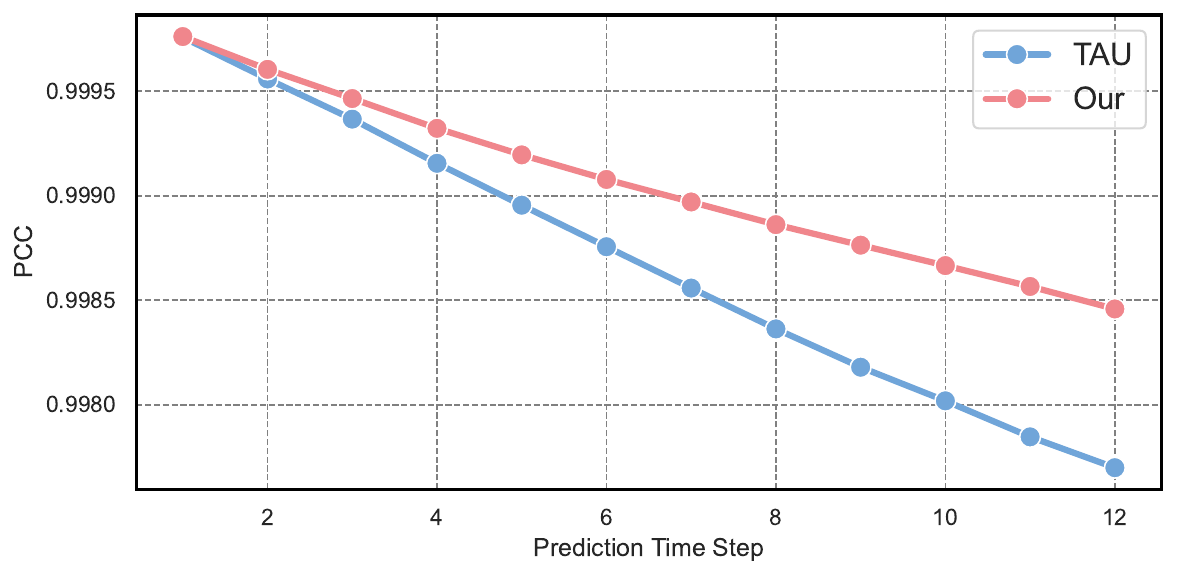} 
    \caption{T2M on PCC.}
  \end{subfigure}
  \hfill
  \begin{subfigure}{0.49\linewidth}
     \includegraphics[width=1\linewidth,  trim=00 00 6 00, clip]{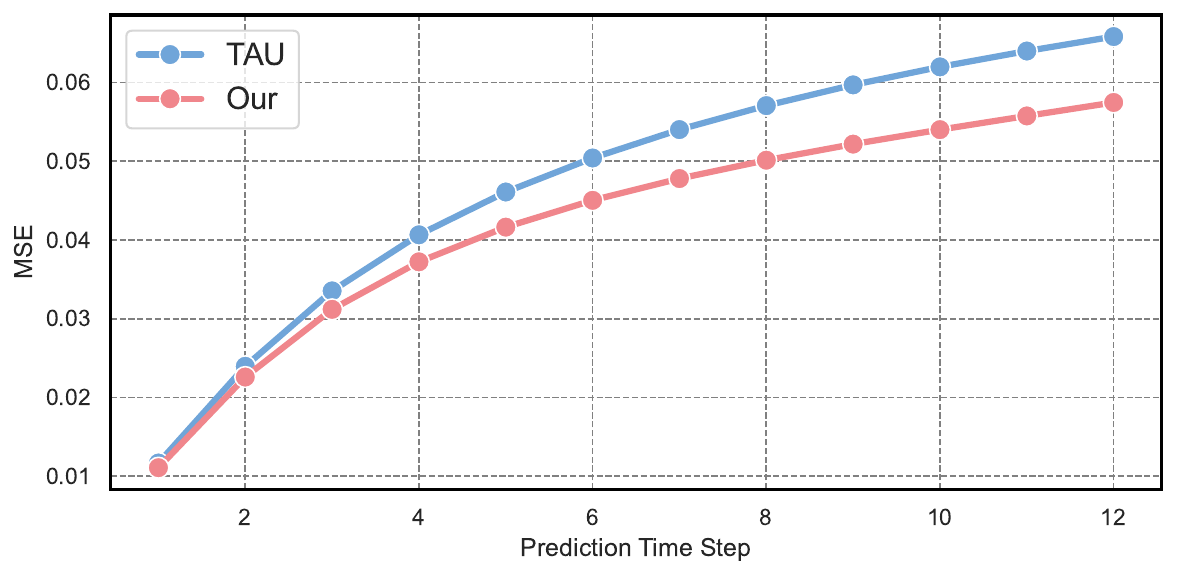} 
    \caption{TCC on MSE.}
  \end{subfigure}
  \hfill
  \begin{subfigure}{0.49\linewidth}
     \includegraphics[width=1\linewidth,  trim=08 00 5 00, clip]{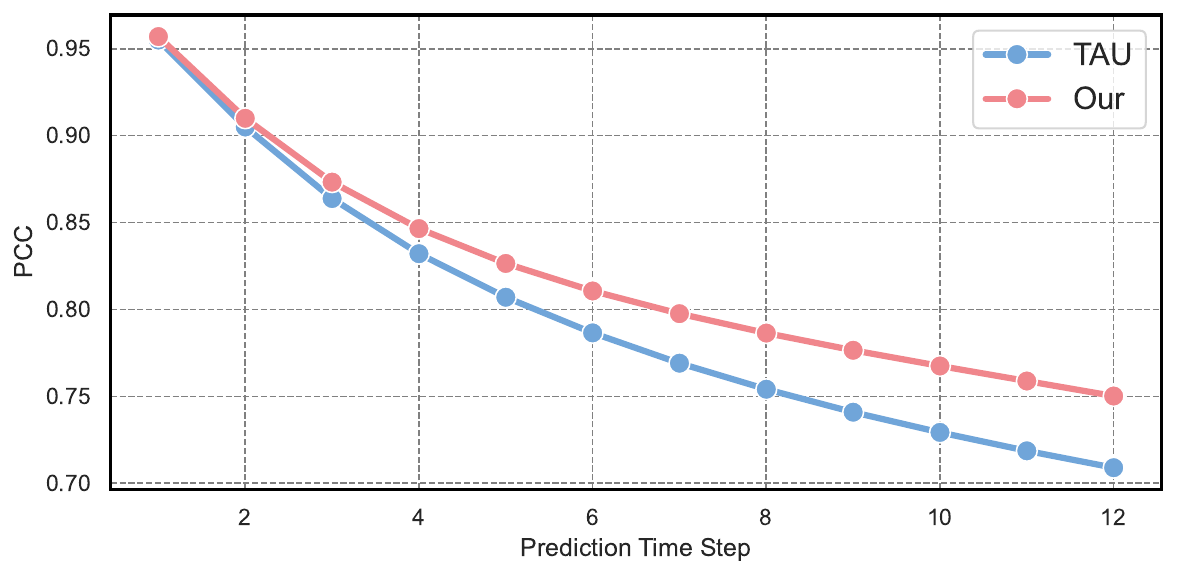} 
    \caption{TCC on PCC.}
  \end{subfigure}
  \caption{Performance comparison of T2M and TCC prediction using MSE and PCC across different prediction time steps.}
  \label{fig:timestep} 
\end{figure}

As shown in Figure~\ref{fig:timestep}, the performance of both models (TAU and our method) varies with the prediction time steps for different variables (T2M and TCC). Although the performance of both models declines as the time step increases, our method consistently outperforms TAU, exhibiting slower growth in MSE and more stable PCC.

\subsection{Single meteorological variables prediction}\label{smvp}
To validate the applicability and effectiveness of our method, we conducted single-variable prediction experiments. Table~\ref{table:uvt2m} presents quantitative comparison results for UV10 and T2M, showing that our method outperforms existing models across all key metrics. Although single-variable accuracy is lower than multi-variable predictions (Table~\ref{table:comp1}), this highlights the effectiveness of our multi-variable fusion approach and the importance of considering multiple variables in meteorological forecasting.
\begin{table}[h]
\small
\centering
\caption{Quantitative comparison on the UV10 and T2M variables. The subscript \( S \) in Baselines indicates the single-variable model.}\vspace{-0.5em}
\begin{tabular}{l|cc|cc}
\hline
\textbf{Method} &  \multicolumn{2}{c|}{UV10} & \multicolumn{2}{c}{T2M} \\

                & MAE & RMSE & MAE & RMSE \\
\hline
ConvLSTM        & 0.9215       & 1.3775        & 0.7949       & 1.2330        \\
PredRNN++       & 0.9019       & 1.3685        & 0.7866       & 1.2070        \\
SimVP           & 0.9510       & 1.4091        & 0.7037       & 1.1130        \\
ConvNeXt         & 0.8698       & 1.3006        & 0.7220       & 1.1300        \\
TAU             & 0.8426       & 1.2619        & 0.6607       & 1.0780        \\
\textbf{Met$^2$Net$_S$}  & 0.8197       & 1.2518        & 0.6536       & 1.0753        \\
\hline
\end{tabular}
\label{table:uvt2m}
\end{table}

\subsection{Additional metrics and resource comparison}
We report both the anomaly correlation coefficient (ACC) and the resource consumption of different models on the cropped ERA5 dataset, as presented in Table~\ref{tab:era5_acc_resource}. All experiments are conducted under the same setting with fp32 precision and batch size 16 on a single NVIDIA RTX 4090 GPU.

Met$^2$Net achieves the highest forecasting accuracy across all variables while maintaining moderate parameter count and competitive efficiency in terms of computation and memory usage.
\begin{table}[ht]
\vspace{-0.5em}
\centering
\caption{ACC and resource comparison on cropped ERA5.}
\vspace{-10pt}
\setlength{\tabcolsep}{0.5mm}
\resizebox{\linewidth}{!}{  
\begin{tabular}{l|cccc|cccc}
\toprule
Method & Params & FLOPs & Mem & Time & MSL & U10 & V10 & T2M \\
 & (M) & (G) & (MiB) & (Min) & \multicolumn{4}{|c}{ACC} \\
\midrule
ConvLSTM & 7.44 & 135.0 & 4398  & 2:42 & 0.9671 & 0.9073 & 0.9427 & 0.9293 \\
MogaNet  & 12.83 & 18.9 & 14408 & 1:37 & 0.9690 & 0.9181 & 0.9492 & 0.9533 \\
TAU      & 12.29 & 18.3 & 11942 & 1:21 & 0.9652 & 0.9097 & 0.9432 & 0.9510 \\
\rowcolor{gray94}\textbf{Met$^2$Net} & 8.90 & 119.0 & 23078 & \textbf{2:24} & \textbf{0.9803} & \textbf{0.9340} & \textbf{0.9590} & \textbf{0.9711} \\
\bottomrule
\end{tabular}
}
\caption*{
\footnotesize Note: All experiments were conducted on a single NVIDIA RTX 4090 GPU. \textbf{Mem} indicates the peak GPU memory usage with fp32 and a batch size of 16; \textbf{Time} refers to the training time per epoch.}
\label{tab:era5_acc_resource}
\vspace{-20pt}
\end{table}

\subsection{Scalability under increased variable input}
To evaluate the scalability of the proposed method, we expand the number of input meteorological variables in the \texttt{$Weather_L$} setting by introducing three additional physical fields: total precipitation (TP), geopotential height (Z), and top-of-atmosphere incoming shortwave radiation (TISR). Correspondingly, we increase the encoder–decoder pairs from 4 to 8, while maintaining the same translator architecture.
We compare the forecasting performance of TAU and Met$^2$Net under varying numbers of encoder–decoder pairs. The results are summarized in Table~\ref{tab:scalability}.
\begin{table}[H]
\centering
\caption{Forecasting performance with increased variables (UV10 and TCC) under different encoder–decoder configurations.}
\label{tab:scalability}
\vspace{-6pt}
\begin{tabular}{l|c|cccc}
\toprule
\textbf{Method}         & \textbf{\# P} & \multicolumn{2}{|c}{\textbf{UV10}} & \multicolumn{2}{c}{\textbf{TCC}} \\ \cline{3-6} 
                        & (M)           & \textbf{MSE}    & \textbf{RMSE}   & \textbf{MSE}   & \textbf{RMSE}   \\ \hline
TAU\textsubscript{1}    & 12.2          & 1.5925          & 1.2619          & 0.0472         & 0.2173          \\
Met$^2$Net\textsubscript{4} & 8.7           & 1.5055          & 1.2270          & 0.0422         & 0.2054          \\
TAU\textsubscript{8}    & 12.2          & 1.7345          & 1.3161          & 0.0444         & 0.2107          \\
\rowcolor{gray94}Met$^2$Net\textsubscript{8} & 8.9 & \bf{1.4740} & \bf{1.2129}  & \bf{0.0417} & \bf{0.2043} \\
\bottomrule
\end{tabular}
\end{table}

The results show that Met$^2$Net maintains superior forecasting accuracy while scaling to more variables, with only a marginal increase in parameter count (from 8.65M to 8.87M). In contrast, TAU exhibits a performance drop despite using the same number of encoders. This demonstrates that Met$^2$Net is well-suited for scalable spatiotemporal modeling in multi-variable settings.

\subsection{Cross-Variable Attention Analysis}
\begin{figure}[H]
    \centering
    \includegraphics[width=0.6\linewidth]{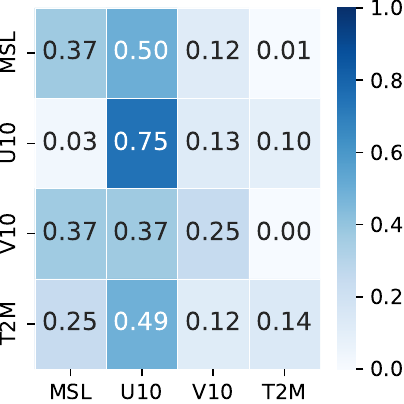}
    \caption{Cross-variable attention map extracted from the translator module. Brighter colors indicate stronger attention. Variables include: MSL, U10, V10, T2M, TP, Z, TISR, and TCC.}
    \label{fig:cross_variable_attention}
\end{figure}

To better understand the inter-variable dependencies captured by the translator module, we analyze the cross-variable attention weights learned during the forecasting process. In our model, each meteorological variable is encoded independently as a token, and the translator performs self-attention over these variable tokens to enable dynamic information aggregation.
Figure~\ref{fig:cross_variable_attention} presents the averaged attention map across all samples and heads. Each row represents the target variable being predicted, and each column indicates the source variable being attended to. The values are normalized attention weights, reflecting how much each variable contributes to the others.
The attention map reveals several meaningful patterns. For example, the model places strong attention between U10 and V10, and between T2M and TCC, which are physically correlated. This indicates that the translator can adaptively capture variable-specific influences, enhancing both forecasting performance and model interpretability.

\subsection{Additional tracking tropical cyclones}\label{app:tropical}
\begin{figure}[h]
  \centering
  \begin{subfigure}{0.49\linewidth}
    \includegraphics[width=1\linewidth]{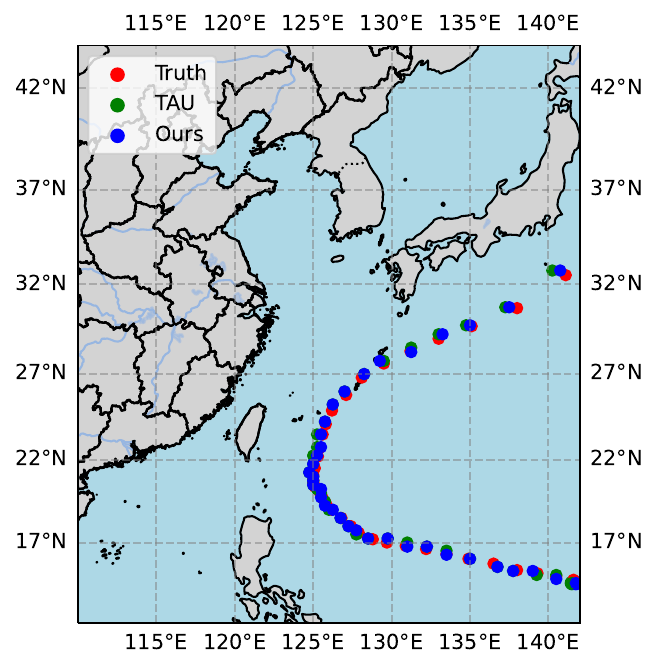} 
    \caption{Predicted and ground truth tracks of Typhoon MAWAR (1-Hour lead time).}
  \end{subfigure}
  \hfill
  \begin{subfigure}{0.49\linewidth}
    \includegraphics[width=1\linewidth]{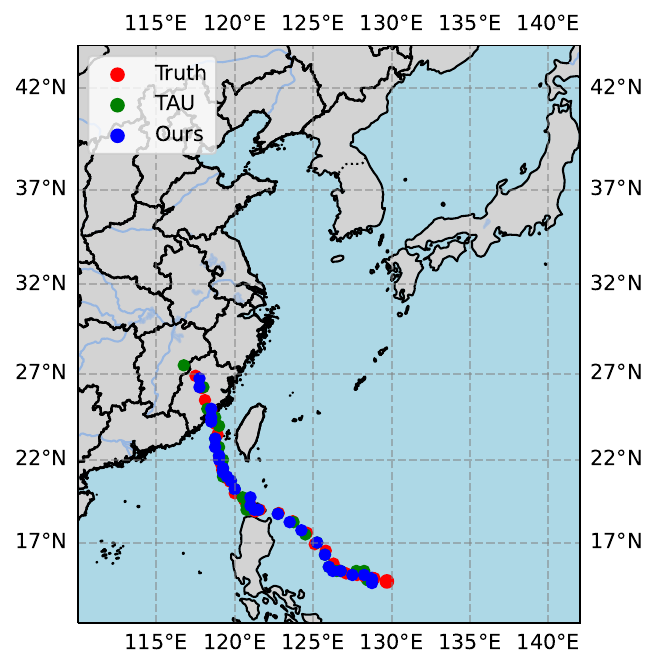} 
    \caption{Predicted and ground truth tracks of Typhoon DOKSURI (3-Hour lead time).}
  \end{subfigure}
  \caption{Tracking tropical cyclones.}
\end{figure}

\subsection{Additional visualization results}\label{app:cases}

\begin{figure*}
  \centering
  \begin{subfigure}{0.49\linewidth}
    \includegraphics[width=1\linewidth]{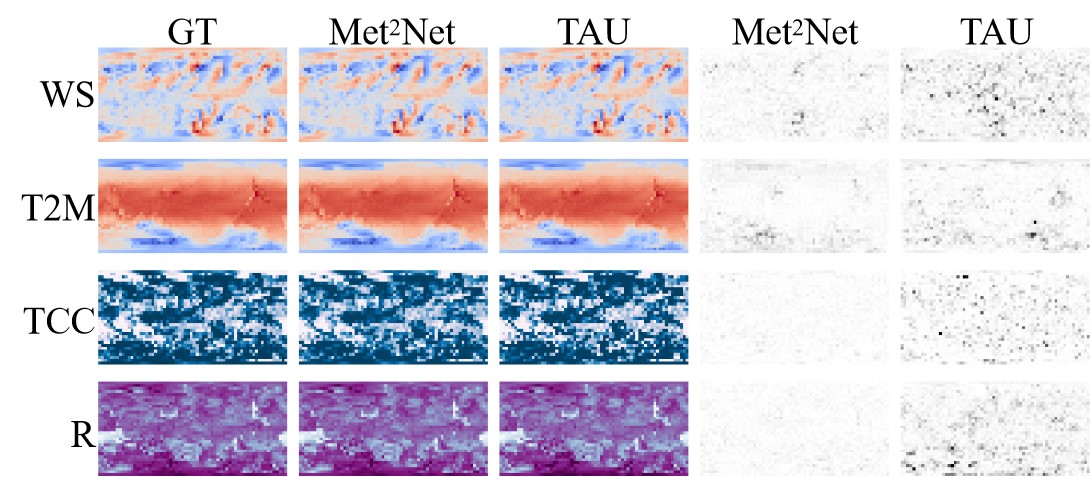} 
    \caption{t=1.}
    \label{fig:app:case1-a}
  \end{subfigure}
  \hfill
  \begin{subfigure}{0.49\linewidth}
   \includegraphics[width=1\linewidth]{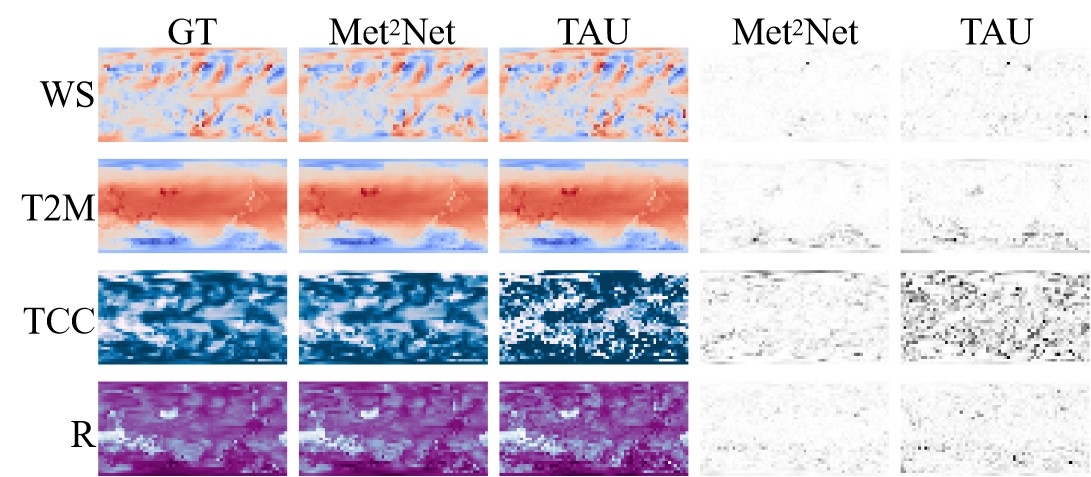} 
    \caption{t=12.}
    \label{fig:app:case1-b}
  \end{subfigure}
  \caption{Visualization of prediction results for different lead times. (a) Results at a forecast time of 1 hour. The background in white represents the absolute error ($|$ GT-Prediction $|$) for each model. (b) Results at a forecast time of 12 hours.}
\end{figure*}

\begin{figure*}
  \centering
  \begin{subfigure}{0.49\linewidth}
    \includegraphics[width=1\linewidth]{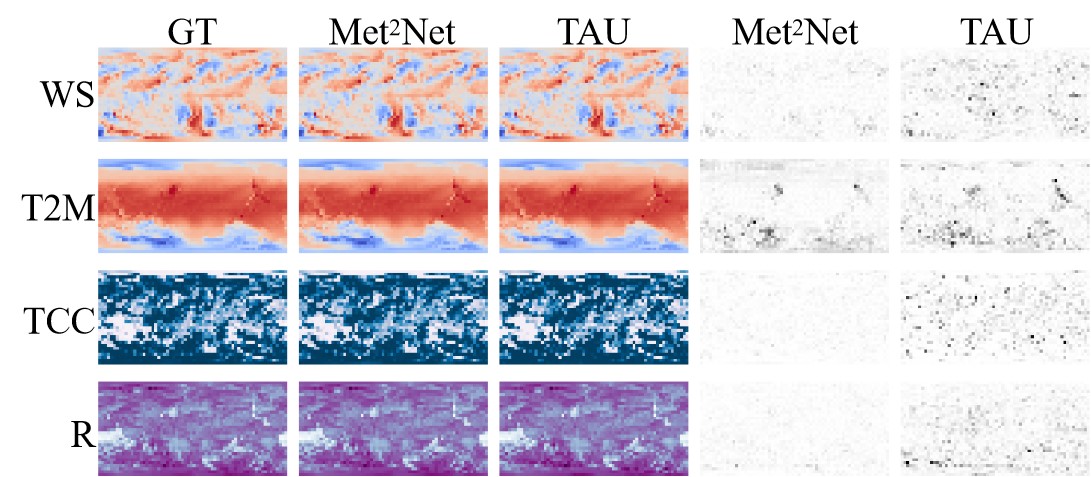} 
    \caption{t=1.}
    \label{fig:app:case2-a}
  \end{subfigure}
  \hfill
  \begin{subfigure}{0.49\linewidth}
   \includegraphics[width=1\linewidth]{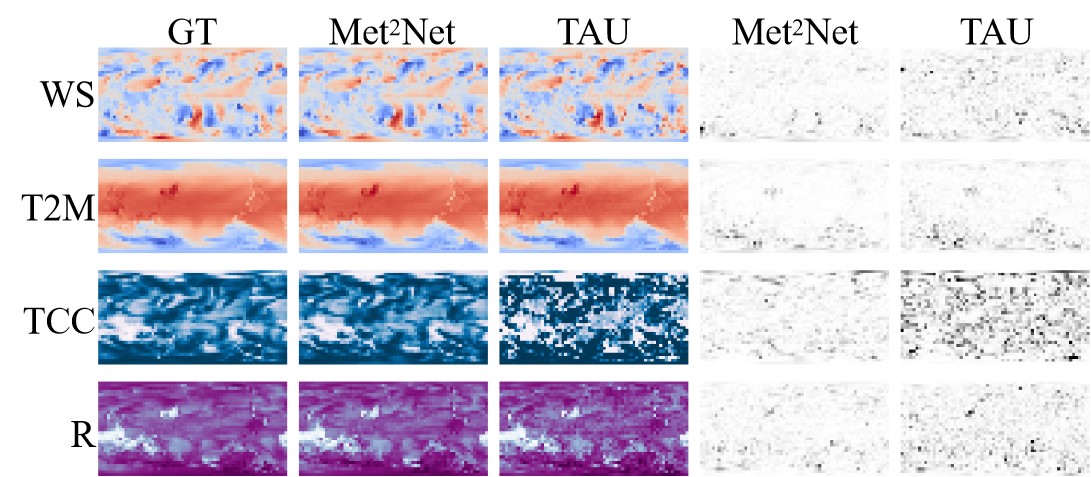} 
    \caption{t=12.}
    \label{fig:app:case2-b}
  \end{subfigure}
  \caption{Visualization of prediction results for different lead times. (a) Results at a forecast time of 1 hour. The background in white represents the absolute error ($|$ GT-Prediction $|$) for each model. (b) Results at a forecast time of 12 hours.}
\end{figure*}

\begin{figure*}
  \centering
  \begin{subfigure}{0.49\linewidth}
    \includegraphics[width=1\linewidth]{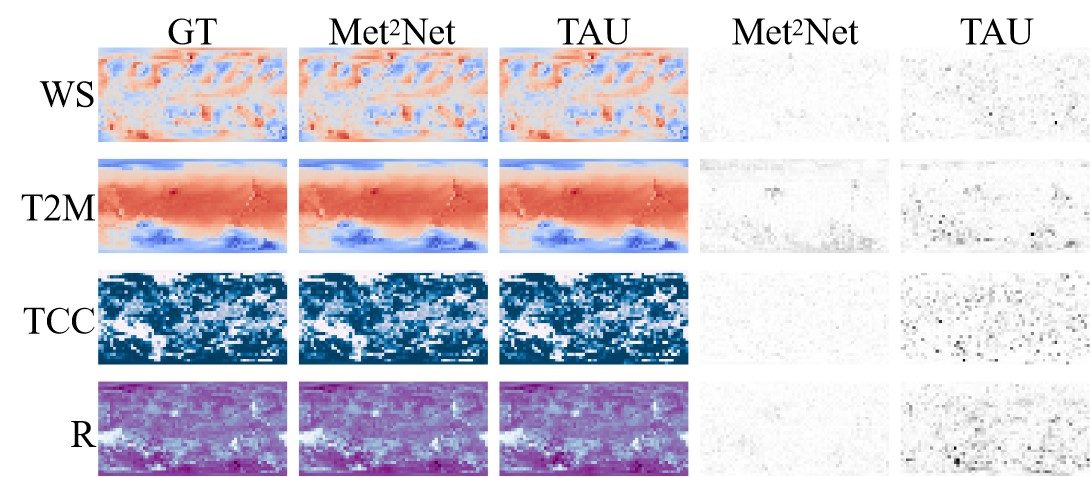} 
    \caption{t=1.}
    \label{fig:app:case3-a}
  \end{subfigure}
  \hfill
  \begin{subfigure}{0.49\linewidth}
   \includegraphics[width=1\linewidth]{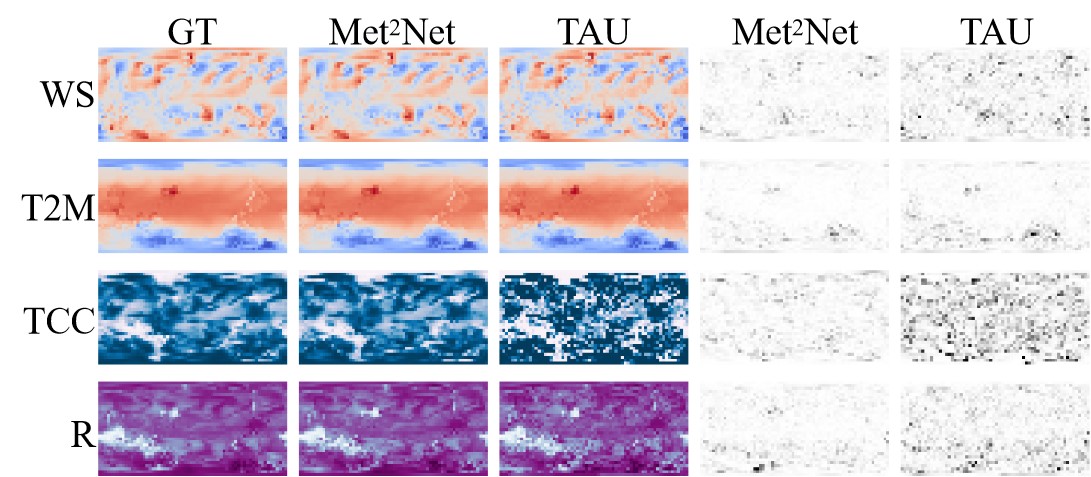} 
    \caption{t=12.}
    \label{fig:app:case3-b}
  \end{subfigure}
  \caption{Visualization of prediction results for different lead times. (a) Results at a forecast time of 1 hour. The background in white represents the absolute error ($|$ GT-Prediction $|$) for each model. (b) Results at a forecast time of 12 hours.}
\end{figure*}

\begin{figure*}
  \centering
  \begin{subfigure}{0.49\linewidth}
    \includegraphics[width=1\linewidth]{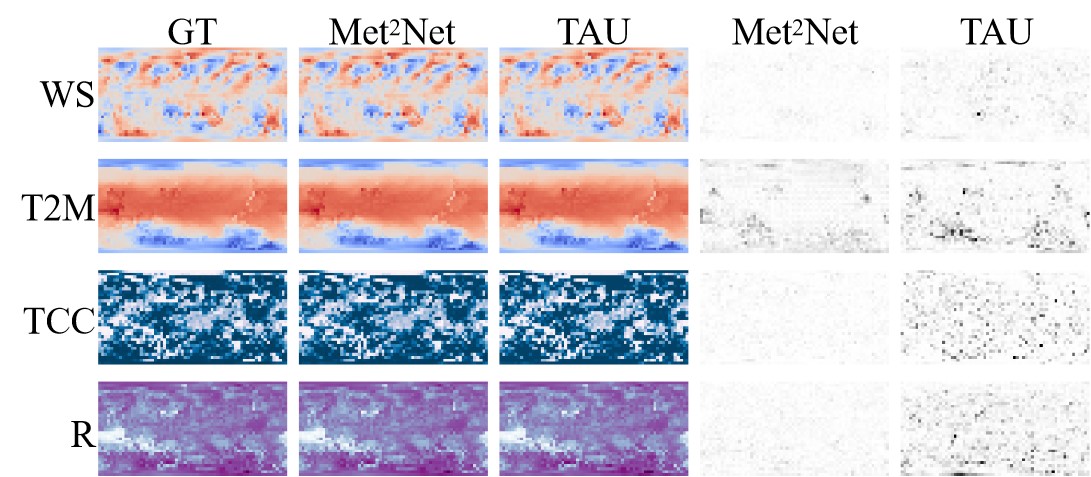} 
    \caption{t=1.}
    \label{fig:app:case4-a}
  \end{subfigure}
  \hfill
  \begin{subfigure}{0.49\linewidth}
   \includegraphics[width=1\linewidth]{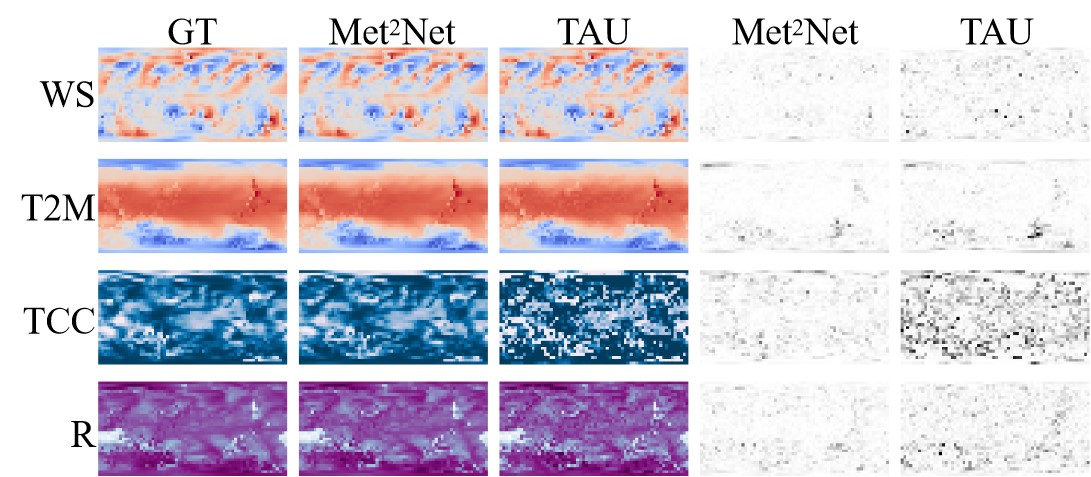} 
    \caption{t=12.}
    \label{fig:app:case4-b}
  \end{subfigure}
  \caption{Visualization of prediction results for different lead times. (a) Results at a forecast time of 1 hour. The background in white represents the absolute error ($|$ GT-Prediction $|$) for each model. (b) Results at a forecast time of 12 hours.}
\end{figure*}

\begin{figure*}
  \centering
  \begin{subfigure}{0.4\linewidth}
    \includegraphics[width=1\linewidth]{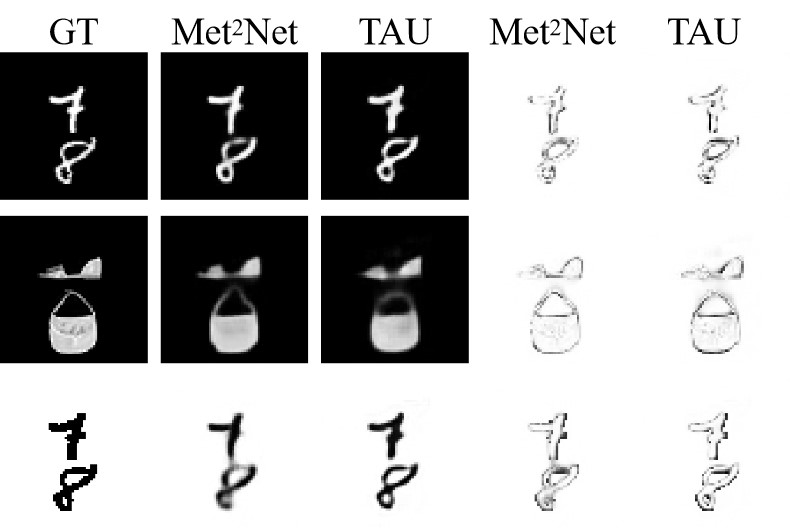} 
    \caption{t=1.}
    \label{fig:app:case5-a}
  \end{subfigure}
  \hspace{0.05\linewidth}
  \begin{subfigure}{0.4\linewidth}
   \includegraphics[width=1\linewidth]{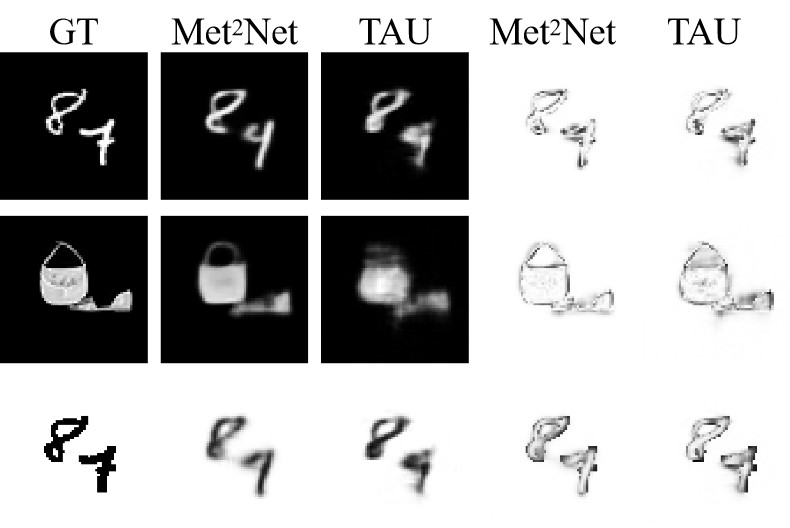} 
    \caption{t=10.}
    \label{fig:app:case5-b}
  \end{subfigure}
  \caption{Visualization of prediction results for different lead times on the Mv$\_$Mmfnist dataset. The last two columns represent the absolute error ($|$ GT - Prediction $|$) for each model. (a) Results at a forecast time of 1 frame. (b) Results at a forecast time of 10 frame.}
\end{figure*}

\begin{figure*}
  \centering
  \begin{subfigure}{0.4\linewidth}
    \includegraphics[width=1\linewidth]{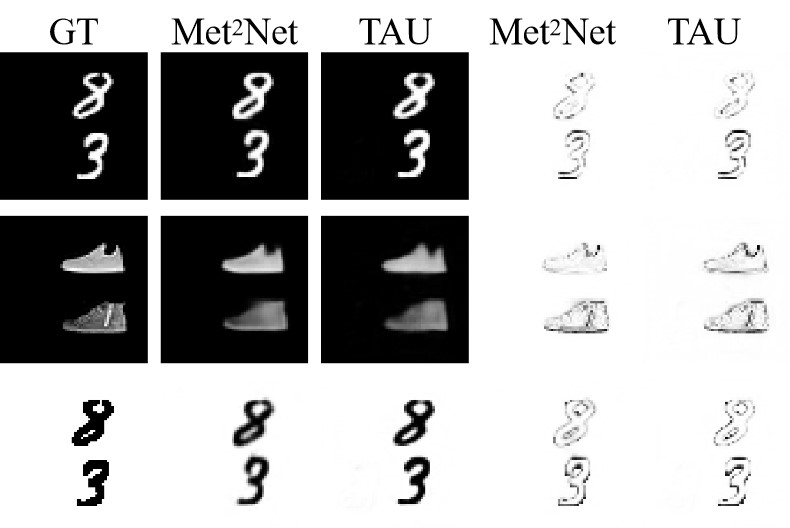} 
    \caption{t=1.}
    \label{fig:app:case6-a}
  \end{subfigure}
  \hspace{0.05\linewidth}
  \begin{subfigure}{0.4\linewidth}
   \includegraphics[width=1\linewidth]{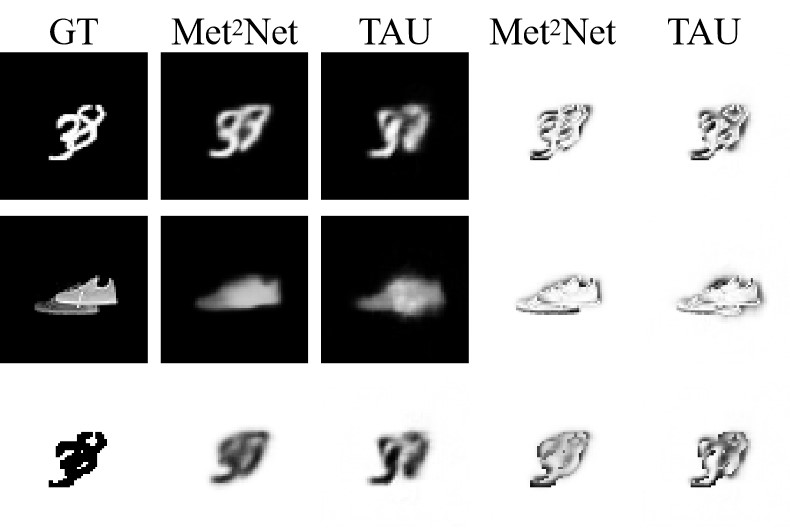} 
    \caption{t=10.}
    \label{fig:app:case6-b}
  \end{subfigure}
  \caption{Visualization of prediction results for different lead times on the Mv$\_$Mmfnist dataset. The last two columns represent the absolute error ($|$ GT - Prediction $|$) for each model. (a) Results at a forecast time of 1 frame. (b) Results at a forecast time of 10 frame.}
\end{figure*}

\begin{figure*}
  \centering
  \begin{subfigure}{0.4\linewidth}
    \includegraphics[width=1\linewidth]{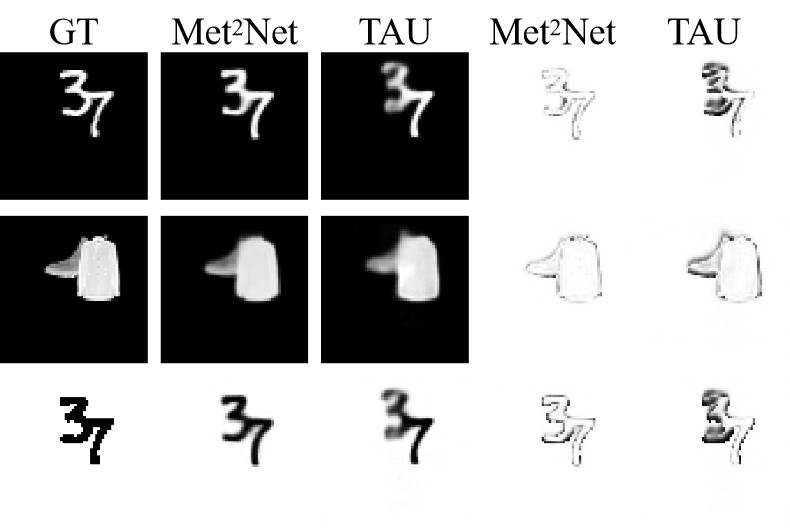} 
    \caption{t=1.}
    \label{fig:app:case7-a}
  \end{subfigure}
  \hspace{0.05\linewidth}
  \begin{subfigure}{0.4\linewidth}
   \includegraphics[width=1\linewidth]{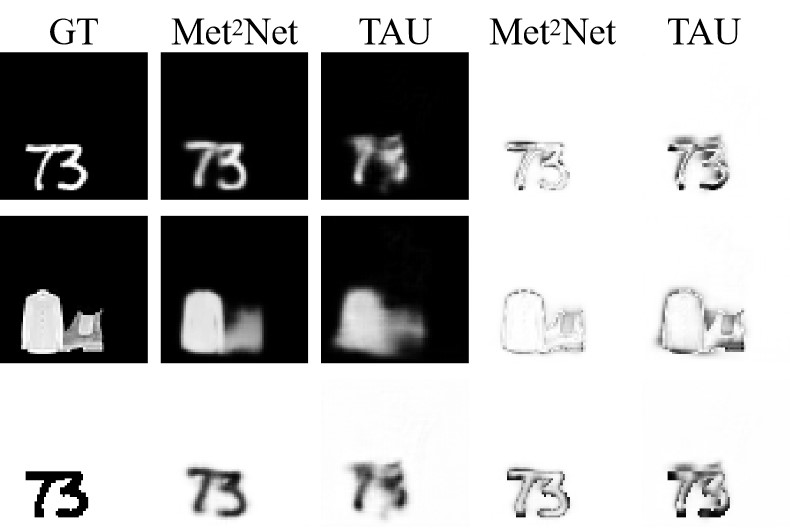} 
    \caption{t=10.}
    \label{fig:app:case7-b}
  \end{subfigure}
  \caption{Visualization of prediction results for different lead times on the Mv$\_$Mmfnist dataset. The last two columns represent the absolute error ($|$ GT - Prediction $|$) for each model. (a) Results at a forecast time of 1 frame. (b) Results at a forecast time of 10 frame.}
\end{figure*}

\end{document}